\documentclass[11pt]{article}

\usepackage[final]{acl}

\usepackage{times}
\usepackage{latexsym}

\usepackage[T1]{fontenc}

\usepackage[utf8]{inputenc}

\usepackage{microtype}

\usepackage{inconsolata}

\usepackage{graphicx}
\usepackage[most]{tcolorbox}

\usepackage{kotex} 
\usepackage{amsmath}
\usepackage{graphicx}
\usepackage{float}
\usepackage{caption}
\usepackage{amssymb}

\usepackage{comment}
\usepackage{hyperref}
\usepackage{booktabs}
\usepackage{tabularx} 
\usepackage{array}
\usepackage{longtable}
\usepackage{ragged2e}
\usepackage{makecell}

\newcolumntype{L}[1]{>{\RaggedRight\arraybackslash}p{#1}}
\newcolumntype{Y}{>{\RaggedRight\arraybackslash}X}

\newcolumntype{Y}{>{\RaggedRight\arraybackslash}X}
\newcolumntype{C}[1]{>{\centering\arraybackslash}p{#1}}

\newtcolorbox{promptbox}[1]{
    breakable,
    enhanced jigsaw,
    width=\linewidth,
    colback=white,
    colframe=black,
    coltitle=white,
    colbacktitle=black,
    title=\textbf{#1},
    fonttitle=\bfseries,
    boxrule=0.8pt,
    arc=2mm,
    left=6pt,
    right=6pt,
    top=5pt,
    bottom=5pt,
    before skip=6pt,
    after skip=6pt
}

\definecolor{caseblue}{HTML}{1F77B4}
\definecolor{casered}{HTML}{D62728}
\definecolor{casegray}{HTML}{F7F7F7}

\newcommand{\cmarkblue}{\textcolor{caseblue}{\ding{51}}}
\newcommand{\xmarkred}{\textcolor{casered}{\ding{55}}}
\newcommand{\goldans}[1]{\textbf{#1}}
\newcommand{\correctans}[1]{\textcolor{caseblue}{\textbf{#1}}}
\newcommand{\wrongans}[1]{\textcolor{casered}{\textbf{#1}}}

\newtcolorbox{qualcasebox}[1]{
    breakable,
    enhanced,
    width=\linewidth,
    colback=white,
    colframe=black!45,
    boxrule=0.55pt,
    arc=1.2mm,
    left=7pt,
    right=7pt,
    top=6pt,
    bottom=6pt,
    before skip=7pt,
    after skip=8pt,
    borderline west={2.2pt}{0pt}{black!65},
    title={\textbf{#1}},
    coltitle=black,
    colbacktitle=gray!10,
    fonttitle=\bfseries
}

\newtcolorbox{maincasebox}{
    breakable,
    enhanced,
    width=\linewidth,
    colback=white,
    colframe=black!45,
    boxrule=0.55pt,
    arc=1.2mm,
    left=7pt,
    right=7pt,
    top=6pt,
    bottom=6pt,
    before skip=7pt,
    after skip=8pt,
    borderline west={2.2pt}{0pt}{black!65}
}

\usepackage[table,xcdraw,dvipsnames]{xcolor}
\usepackage{colortbl}
\usepackage{multirow}
\usepackage{makecell}
\usepackage{arydshln}

\tcbuselibrary{breakable}
\usepackage[normalem]{ulem}
\useunder{\uline}{\ul}{}
\usepackage{enumitem}
\usepackage{pifont}

\usepackage{algorithm}
\usepackage{algorithmic}

\usepackage{hyperref}
\usepackage{titletoc}

\usepackage{cuted}
\usepackage{placeins}

\usepackage{amsmath}
\title{ESGPT: A Data-Centric ESG Language Model with Unified Instruction and Benchmark Resources}

\author{
  \textbf{Jaeyoung Lee\textsuperscript{1}}
  \quad
  \textbf{Misuk Kim\textsuperscript{1,2,*}}
\\
  \textsuperscript{1}Department of Data Science, Hanyang University, Seoul, Republic of Korea
\\
  \textsuperscript{2}Department of Artificial Intelligence, Hanyang University, Seoul, Republic of Korea
\\
  \texttt{\{jaeylee, misukkim\}@hanyang.ac.kr}
\\
\textbf{GitHub:} \url{https://github.com/nowzer0/ESGPT}
}

\begin{document}
\maketitle

\renewcommand{\thefootnote}{}
\footnote{$^*$Corresponding author.}

\begin{abstract}
ESG analysis requires complex language understanding across diverse sources, including corporate reports, news, and disclosures, for tasks such as ESG relevance detection, action interpretation, impact analysis, and question answering. However, existing ESG datasets are fragmented by task and source, with heterogeneous formats and annotation schemes, limiting their use for training general-purpose ESG LLMs and unified evaluation. In this work, we construct \textsf{ESG-Instruct}, a 30,000-instance ESG instruction dataset built from public ESG resources, and \textsf{ESGU-Bench}, an ESG understanding benchmark covering six representative ESG tasks. Based on these resources, we further develop \textsf{ESGPT}, an 8B ESG-specialized LLM, and evaluate it against diverse open-source and API-based LLMs. Experimental results show that general-purpose LLMs exhibit substantial performance variation across ESG tasks, while \textsf{ESGPT}, despite its relatively small scale, achieves the best performance on multiple ESG tasks. This work provides unified data and benchmark resources for training and evaluating ESG-specialized LLMs.

\end{abstract}

\section{Introduction}
\begin{figure*}[t!]
    \centering
    \includegraphics[width=1\linewidth]{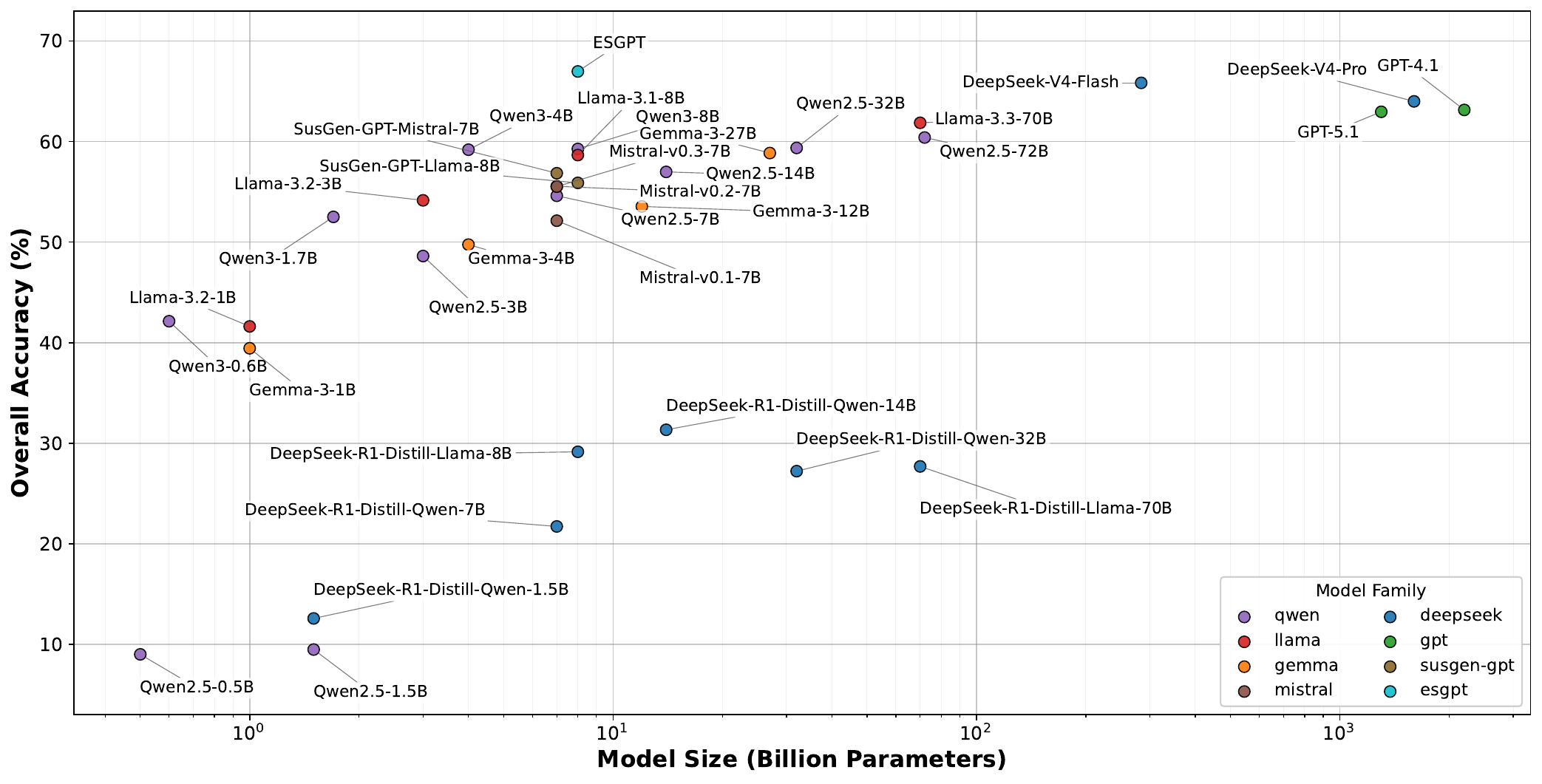}
    \caption{Overall accuracy by model size on the \textsf{ESGU-Bench} across 35 LLMs.}
    \label{fig:overall_acc}
\end{figure*}
Environmental, Social, and Governance (ESG) has become a key criterion for assessing corporate sustainability and social responsibility. ESG information appears in diverse forms, including sustainability reports, disclosures, news articles, and international standards, and requires complex understanding and reasoning beyond simple classification, such as relevance detection, claim interpretation, impact analysis, and question answering. Large Language Models (LLMs) have shown the potential to address these ESG tasks through natural language interfaces~\cite{sun2026esg}. However, as ESG is a specialized domain whose interpretation depends on industry context, disclosure standards, and supporting evidence, strong performance on general benchmarks does not necessarily indicate ESG understanding. Therefore, ESG-specific data and benchmarks are needed to evaluate ESG understanding capabilities~\cite{he2025esgenius}.

Despite this need, publicly available ESG resources remain limited~\cite{ong2025towards, lee2023esginfo, lee2024esgkor}. Existing datasets are fragmented across languages, input structures, output formats, annotation units, and evaluation criteria, limiting their use for training general-purpose ESG LLMs and for unified evaluation. Moreover, ESG data construction often requires access to restricted source materials and substantial domain expertise. This highlights the need to systematically leverage high-quality public ESG resources and build an integrated data foundation for consistent training and evaluation across diverse ESG tasks.
This work constructs \textsf{ESG-Instruct}, a large-scale ESG instruction dataset, by systematically selecting and processing public ESG data resources. Based on this dataset, we develop \textsf{ESGPT}, a data-centric ESG LLM, and propose \textsf{ESGU-Bench}~(ESG Understanding Benchmark), which is based on human annotation and expert review. Experiments on 35 LLMs ranging from 0.5B to 2.2T parameters show that general-purpose LLMs exhibit substantial variation across ESG tasks, and even some large API-based LLMs show limited performance. In contrast, \textsf{ESGPT} achieves strong performance on multiple ESG tasks despite its relatively small scale. The contributions of this work are as follows:

\begin{itemize}
    \item We construct \textsf{ESG-Instruct}, a large-scale ESG instruction dataset, by systematically selecting and integrating public ESG data resources, enabling diverse ESG understanding tasks to be learned in a consistent instruction format.
    \item We construct \textsf{ESGU-Bench}, consisting of six representative tasks centered on human-annotated and expert-reviewed datasets, to reliably evaluate the ESG understanding capability of LLMs.
    \item We develop \textsf{ESGPT}, a data-centric ESG LLM based on \textsf{ESG-Instruct}, and analyze the limitations of general-purpose LLMs and the effectiveness of ESG-specialized models through large-scale comparative experiments.
\end{itemize}


\section{Related Work}
Datasets and benchmarks for ESG analysis have been built from diverse sources, including corporate reports, news articles, disclosures, and climate documents. Early studies focused mainly on classification tasks such as ESG relevance detection, topic classification, and sentiment analysis, while recent work has expanded to more complex ESG understanding tasks, including corporate action interpretation, claim verification, and question answering. ESG-FTSE~\cite{pavlova2024esgftse} and ESG-Kor~\cite{lee2024esgkor} address ESG sentence detection and classification, whereas A3CG~\cite{ong2025towards} and PromiseEval~\cite{seki2025mlpromise} target fine-grained ESG interpretation of claims, actions, promises, and evidence. SustainableQA~\cite{ali2025sustainableqa} and ESGenius~\cite{he2025esgenius} further provide QA-based benchmarks for evaluating ESG and sustainability knowledge.

From the model perspective, domain-specific pre-trained language models (PLMs), such as ESGBERT~\cite{schimanski2024ESGBERT} and ClimateBERT~\cite{webersinke2021climatebert}, have been used for ESG and climate-related text analysis. More recently, research has expanded toward LLM- or retrieval-augmented generation-based ESG analysis, question answering, and report generation, as in ChatReport~\cite{ni2023chatreport}, ESGReveal~\cite{zou2025esgreveal}, and SusGen-GPT~\cite{wu2025susgen}.

However, existing studies mainly focus on individual tasks, specific sources, or particular applications. As a result, integrated instruction datasets for general-purpose ESG LLM development and reliable benchmarks for ESG understanding remain limited. Therefore, this work jointly presents an ESG instruction dataset, an ESG understanding benchmark, and an ESG-specialized LLM by consolidating and reformulating public ESG resources.

\section{ESG Resource \& Model Construction}
\label{sec:construction}
This study collects and reorganizes public ESG datasets to build integrated resources for training and evaluating ESG-specialized LLMs. The data are organized by task objective, language, source type, annotation scheme, answer structure, and output requirements, followed by quality control to identify duplicates, missing values, formatting errors, label inconsistencies, and task-unsuitable instances. While preserving the original task definitions and answer structures, we convert the data into chat-prompt-based instances through translation, format normalization, task refinement, and prompt design. The refined tasks are then reorganized from the perspective of ESG understanding capabilities and integrated into six high-level categories. Dataset construction details are provided in Appendix~\ref{app:esg_resources}.

\begin{figure}[t]
    \centering
    \includegraphics[width=1\linewidth]{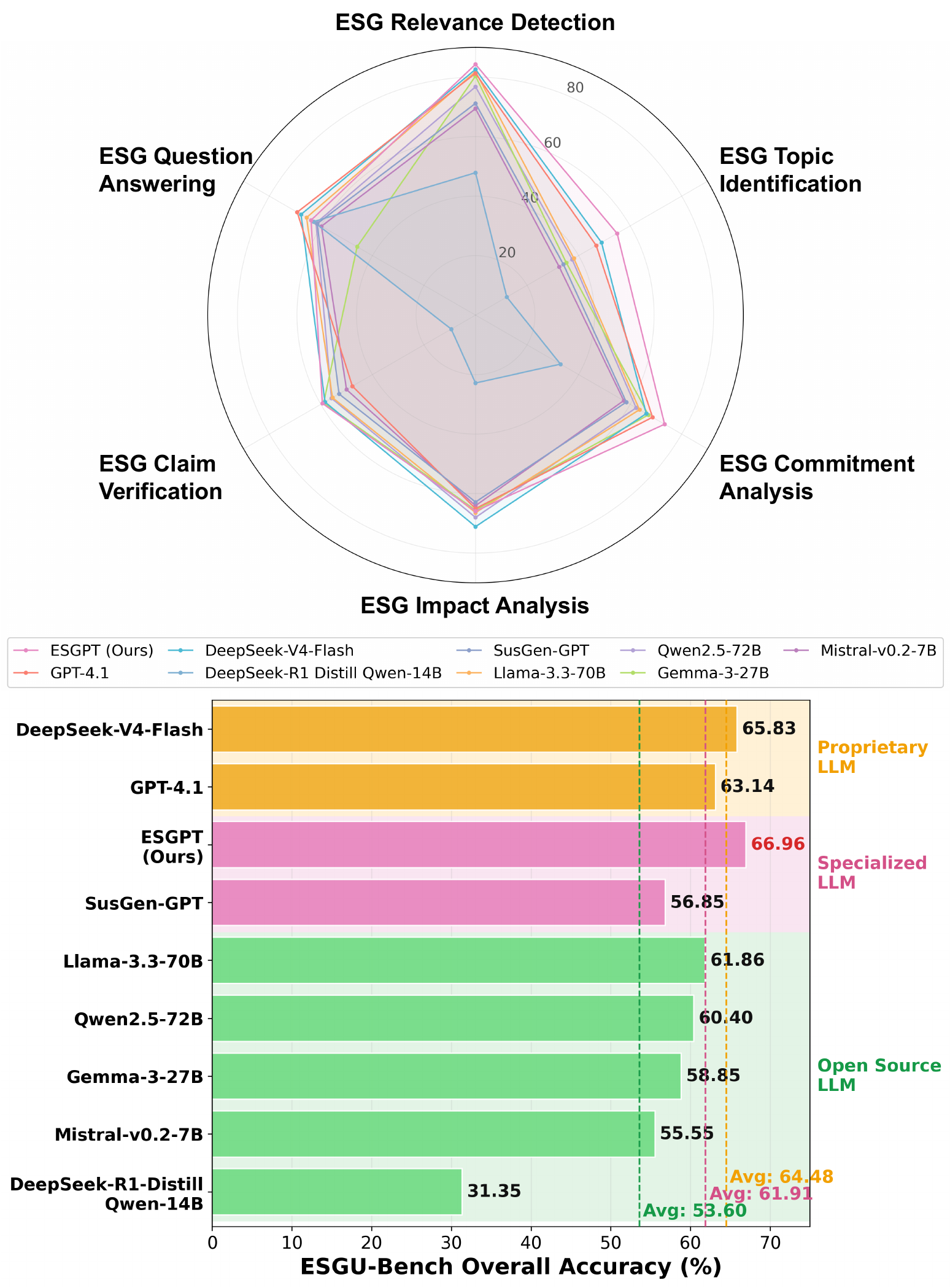}
    \caption{Main results on \textsf{ESGU-Bench}: category-level accuracy (top) and overall accuracy of LLMs (bottom). Dashed vertical lines indicate the average overall accuracy of each model group.}
    \label{fig:main_results}
\end{figure}
\paragraph{\textsf{ESG-Instruct}.}
\textsf{ESG-Instruct} is a 30,000-instance instruction dataset for training ESG-specialized LLMs. From quality-reviewed ESG data, we construct 35 instruction tasks and select training instances by considering task coverage and data balance~\cite{wu2025susgen}. We adjust data proportions across the six ESG understanding categories to mitigate bias toward specific tasks or datasets. To support both ESG domain adaptation and general instruction-following ability, we allocate 10\% of the dataset to general instruction data. Considering the limited availability of public ESG resources, we also include translated data and a small amount of LLM-assisted data.


\paragraph{\textsf{ESGU-Bench}.}
\textsf{ESGU-Bench} is an ESG understanding benchmark composed of human-annotated and expert-reviewed datasets. It includes 9,514 instances and 21 evaluation tasks, which are integrated into six high-level ESG understanding categories to evaluate diverse ESG understanding capabilities beyond single classification- or QA-centered evaluation.

\paragraph{\textsf{ESGPT}.}
\textsf{ESGPT} is a data-centric ESG LLM trained on \textsf{ESG-Instruct}. Rather than introducing a new model architecture, this study focuses on analyzing how the design of ESG instruction data, considering task coverage, source diversity, label balance, and data quality, affects ESG understanding performance.

\begin{figure}[t]
    \centering
    \includegraphics[width=1\linewidth]{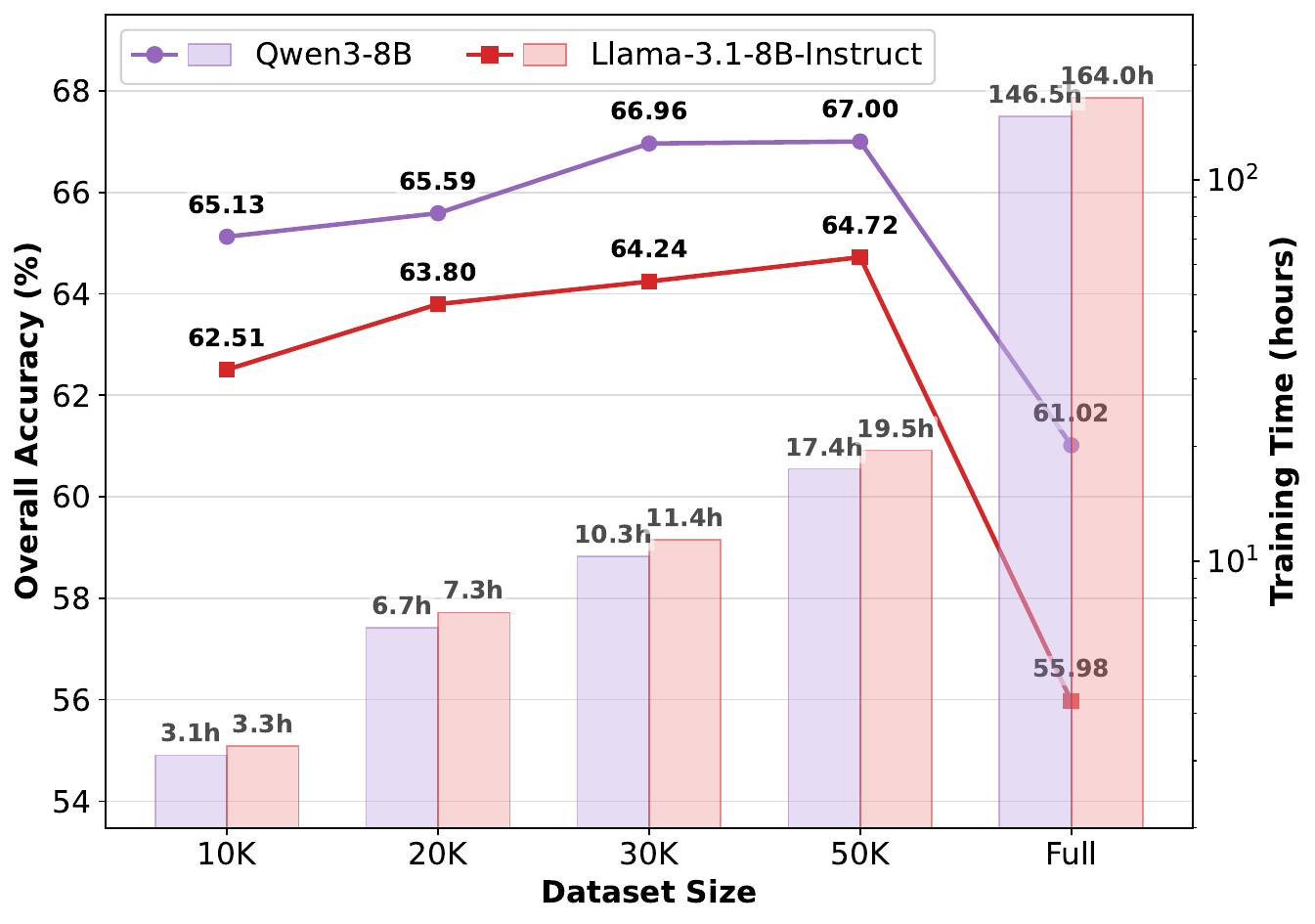}
    \caption{Effect of \textsf{ESG-Instruct} data scale on \textsf{ESGPT}. Lines denote overall accuracy on \textsf{ESGU-Bench}, and bars denote training time in hours. All data-scale variants were trained with the same fixed random seed and training configuration and evaluated under the same protocol.}
    \label{fig:ablation}
\end{figure}

\section{Experiments}
\label{sec:experimental_setup}
Using \textsf{ESGU-Bench}, we conduct large-scale comparative experiments on 35 LLMs of diverse types and scales. All models are evaluated in a zero-shot setting, and performance is reported in terms of accuracy for each of the six ESG understanding categories and overall accuracy. All experiments were conducted on a single NVIDIA A100 GPU. Training and experimental details are provided in Appendix~\ref{app:implementation}, and detailed results and analyses are provided in Appendix~\ref{app:additional_results}.

\paragraph{\textsf{ESGU-Bench} Results.}
Figure~\ref{fig:main_results} shows the performance of representative LLMs on \textsf{ESGU-Bench}. In the main text, we report results focusing on the model with the highest overall accuracy within each model family. ESG understanding performance varies substantially across model families and task types, and even large proprietary LLMs do not show consistent advantages across all ESG categories.
In contrast, \textsf{ESGPT}, despite being an 8B model, achieves the best performance in four of the six categories and outperforms both large API-based models from the DeepSeek and GPT families and open-source ESG LLMs, with an overall accuracy of 66.96\%. This demonstrates that systematically constructed ESG instruction data alone can lead to strong ESG understanding performance. At the same time, performance limitations are still observed in some ESG tasks, suggesting the need for more refined ESG contextual understanding and domain-specific reasoning. 

We also qualitatively analyze the ESG understanding performance of \textsf{ESGPT} and general-purpose LLMs. Input prompts and inputs are presented in summarized form.
\begin{maincasebox}
\small
\textbf{[User]} 
Classify the ESG impact severity. \newline Allowed labels: Low, Medium, High.
\vspace{0.45em}

\textbf{[Input]} A.P. Møller acquires Faerch Group, a sustainable food packaging company providing circular food packaging solutions, products made from up to 100\% recycled content, fully recyclable food-grade packaging, and an integrated recycling system.
\tcblower
\scriptsize
\textbf{[Gold and Model Answers]}
\vspace{0.15em}

\setlength{\tabcolsep}{2.5pt}
\renewcommand{\arraystretch}{1.02}
\begin{tabularx}{1.0\linewidth}{@{}Y C{0.30\linewidth} C{0.30\linewidth}@{}}
\hline
\textbf{Model} & \textbf{Answer} & \textbf{Result} \\
\hline
Gold & \goldans{High} & -- \\
\hline
DeepSeek & \wrongans{Medium} & \xmarkred \\
GPT & \wrongans{Low} & \xmarkred \\
\textsf{ESGPT} & \correctans{High} & \cmarkblue \\
\hline
\end{tabularx}
\end{maincasebox}
\textsf{ESGPT} predicts \texttt{High} by recognizing the acquisition of Faerch Group, with its \textit{circular packaging and integrated recycling system}, as a \textit{strategically significant ESG event}. In contrast, the DeepSeek and GPT-family models appear to treat the case as \textit{a general ESG-related corporate activity} and assign a lower impact severity. This suggests that an ESG-specialized LLM can better capture the \textit{ESG significance and impact severity of corporate events}. 
Additional cases and detailed analyses are provided in Appendix~\ref{app:case_studies}.


\paragraph{\textsf{ESG-Instruct} Scale Analysis.}
Figure~\ref{fig:ablation} shows performance changes according to the scale of \textsf{ESG-Instruct}. With only 10K training instances, the model already ranks among the top-performing models, and further improves when scaled to 30K. This trend is consistent across different backbone models. By contrast, the full-scale setting degrades performance despite requiring the highest training cost. These results suggest that task/category balance and data quality are more important for ESG instruction tuning than simply increasing data scale. In particular, tasks with limited high-quality annotations are difficult to retain at sufficient proportions in the full-scale setting, while data-rich tasks may become overrepresented.

\begin{table}[t]
\centering
\small
\setlength{\tabcolsep}{3pt}
\renewcommand{\arraystretch}{1.12}
\begin{tabularx}{\linewidth}{@{}XX@{}}
\hline
\textbf{\textsf{ESGPT}} & \textbf{SusGen-GPT} \\
\hline
The board indirectly oversees climate-related risks through operational safety and financial leverage reduction, but \textcolor{caseblue}{the report does not explicitly link these strategies to climate-specific measures}. 
&
The board oversees climate-related risks through governance mechanisms, and the Petrobras Corruption Prevention Program \textcolor{casered}{likely includes climate-related risks}. \\
\hline
\end{tabularx}
\caption{Compressed example of TCFD-style report generation.}
\label{tab:main_report_generation_case}
\end{table}

\paragraph{Report Generation Analysis.}
We qualitatively compare the TCFD~(Task Force on Climate-related Financial Disclosures) report generation behavior of \textsf{ESGPT} and SusGen-GPT, a sustainability report generation model (Table~\ref{tab:main_report_generation_case}). A representative case is presented in the main text, with the full comparison in Appendix~\ref{app:report_generation}.

\textsf{ESGPT} notes that operational safety and financial strategy may be indirectly related to climate risks, but does not infer climate-specific oversight without explicit evidence. In contrast, SusGen-GPT links even the Petrobras Corruption Prevention Program to climate-related risk oversight, extending general governance information into climate-specific disclosure. Since ESG report generation is tied to corporate disclosure, such unsupported extensions can undermine reliability. This case suggests that \textsf{ESGPT} better preserves the boundary between general ESG information and climate-specific disclosure, supporting more reliable and evidence-grounded ESG report generation.



\section{Conclusion}
\label{sec:conclusion}
This study proposed \textsf{ESG-Instruct}, \textsf{ESGU-Bench}, and \textsf{ESGPT}, a data-centric ESG LLM, for ESG-specialized LLM development and reliable evaluation. These resources are designed to support consistent training and evaluation across diverse ESG understanding tasks. Experiments show that general-purpose LLMs and large API-based models exhibit substantial performance variation across ESG tasks, whereas \textsf{ESGPT} achieves strong ESG understanding performance despite its relatively small scale. These results demonstrate the effectiveness of the proposed ESG domain-specific instruction data and integrated benchmark for training and evaluating ESG-specialized LLMs.

\section*{Limitations}
Although this study proposes an integrated benchmark for ESG understanding and an ESG-specialized instruction model, it has several limitations. 
First, while \textsf{ESGU-Bench} covers diverse ESG understanding tasks, including ESG relevance detection, topic identification, commitment analysis, impact analysis, claim verification, and question answering, it cannot fully represent all ESG disclosure contexts across industries, regions, and regulatory frameworks. 
ESG issues can be interpreted differently depending on industry-specific materiality and country-specific disclosure requirements. 
Nevertheless, we collected a wide range of publicly available ESG datasets and constructed the benchmark while considering the distribution of tasks and data sources, in order to maintain broad coverage of ESG understanding as much as possible.

Second, the ESG data used in this study is largely English-centric. 
This reflects the fact that many publicly available ESG reports, news articles, standard documents, and benchmark datasets are primarily constructed in English. 
We translated some non-English ESG data into English and used them for training; however, regional expressions, institutional nuances, and disclosure practices in the original texts may have been partially weakened during translation. 
Nevertheless, during the translation process, we sought to preserve the original meaning as much as possible by retaining numerical values, proper nouns, commitment expressions, verification-related wording, paragraph structure, and the domain-specific context of ESG, sustainability, finance, and disclosure. Future work should extend \textsf{ESG-Instruct} and \textsf{ESGU-Bench} toward multilingual and jurisdiction-aware ESG resources, and further develop \textsf{ESGPT} to better reflect such linguistic and regional diversity. Since high-quality ESG resources remain limited for non-English languages and region-specific disclosure contexts, future ESG-specialized datasets, benchmarks, and models should better capture differences in regulatory environments, regional reporting practices, institutional terminology, and local ESG expressions.

Third, due to training resource constraints, \textsf{ESGPT} was developed by applying LoRA-based training to an 8B-scale base model. 
This setting enables ESG-specialized adaptation at a relatively limited training cost, but this study does not fully examine how performance may change with larger-scale models or full fine-tuning.
Our goal is not to exhaustively optimize architecture scaling, but to provide \textsf{ESG-Instruct}, \textsf{ESGU-Bench}, and \textsf{ESGPT} as a unified foundation for ESG-specialized LLM research and to demonstrate the feasibility of resource-efficient ESG adaptation.
Since ESG understanding requires fine-grained label boundary recognition, disclosure-context understanding, and evidence-based judgment, the impact of model scale and training strategy should be further investigated in future work.

Finally, ESG-specialized models may produce incorrect or incomplete judgments, especially in cases that require industry-specific expertise, regulatory interpretation, or evidence-based verification. Therefore, \textsf{ESGPT} and \textsf{ESGU-Bench} should be used as decision-support resources rather than substitutes for expert review in corporate disclosure, investment, or regulatory decision-making. 
Despite these limitations, this study provides a systematic benchmark for ESG understanding and demonstrates the potential of ESG-specialized LLMs, offering a foundation for developing more reliable ESG AI systems.

\section*{Ethical Considerations}
This study uses publicly available ESG datasets and does not involve human subject experiments or personal information. The original dataset creators and sources are credited in Appendix~\ref{app:esg_resources}. Therefore, this work does not raise additional ethical concerns related to human subjects research or personal data processing.

\section*{Acknowledgments}
This work was supported by the Institute of Information \& Communications Technology Planning \& Evaluation (IITP) grant funded by the Korea government (MSIT) (No. RS-2020-II201373, Artificial Intelligence Graduate School Program (Hanyang University)) and the Institute of Information \& Communications Technology Planning \& Evaluation (IITP) grant funded by the Korea government (MSIT, Ministry of Science and ICT) (No. RS-2026-25548925).

\bibliography{reference}

\clearpage
\appendix
\setcounter{table}{0}
\setcounter{figure}{0}
\setcounter{equation}{0}

\renewcommand{\thetable}{A.\arabic{table}}
\renewcommand{\thefigure}{A.\arabic{figure}}

\appendix

\section{ESG Resources}
\label{app:esg_resources}
We describe the collection and construction process of the ESG data resources used in this study. We collected ESG datasets with clear provenance and reproducibility for training and evaluating ESG-specialized LLMs. Datasets were selected primarily from public repositories such as Hugging Face, GitHub, and Kaggle, or from resources whose construction process, annotation scheme, label system, and source materials could be verified through papers or official documents. Some datasets were also obtained by directly requesting them from the original authors or providers. In such cases, we included them only when their provenance, task definition, annotation scheme, and answer format were verifiable. This process excludes arbitrarily collected or unofficial data whose provenance or answer structure cannot be verified.

The collected datasets cover diverse ESG-related sources, including corporate sustainability reports, climate disclosures, news and media articles, and ESG standards and frameworks. For each dataset, we manually reviewed its task objective, language, source type, annotation scheme, label system, answer format, and output requirements to assess its suitability as a training or evaluation resource. Table~\ref{tab:esg_resources} presents the datasets used in this study and their related information.

Since the collected datasets have different input structures and output formats, we applied a common preprocessing process for LLM training and evaluation. Specifically, we removed duplicate samples, checked missing fields, corrected formatting errors, normalized label names, unified answer formats, and reviewed task suitability. When a dataset contained multiple subtasks or hierarchical labels, we refined the task units so that each task objective and output requirement were clearly specified.

Through this process, we constructed two ESG data resources. First, \textsf{ESG-Instruct} is an instruction-tuning dataset for training ESG-specialized LLMs, aiming to support domain adaptation across diverse ESG understanding tasks. Second, \textsf{ESGU-Bench} is a benchmark for reliably evaluating the ESG understanding capability of LLMs, centered on evaluation data with human annotations, expert review, or clear gold answers.



\begin{figure}[h]
    \centering
    \includegraphics[width=1\linewidth]{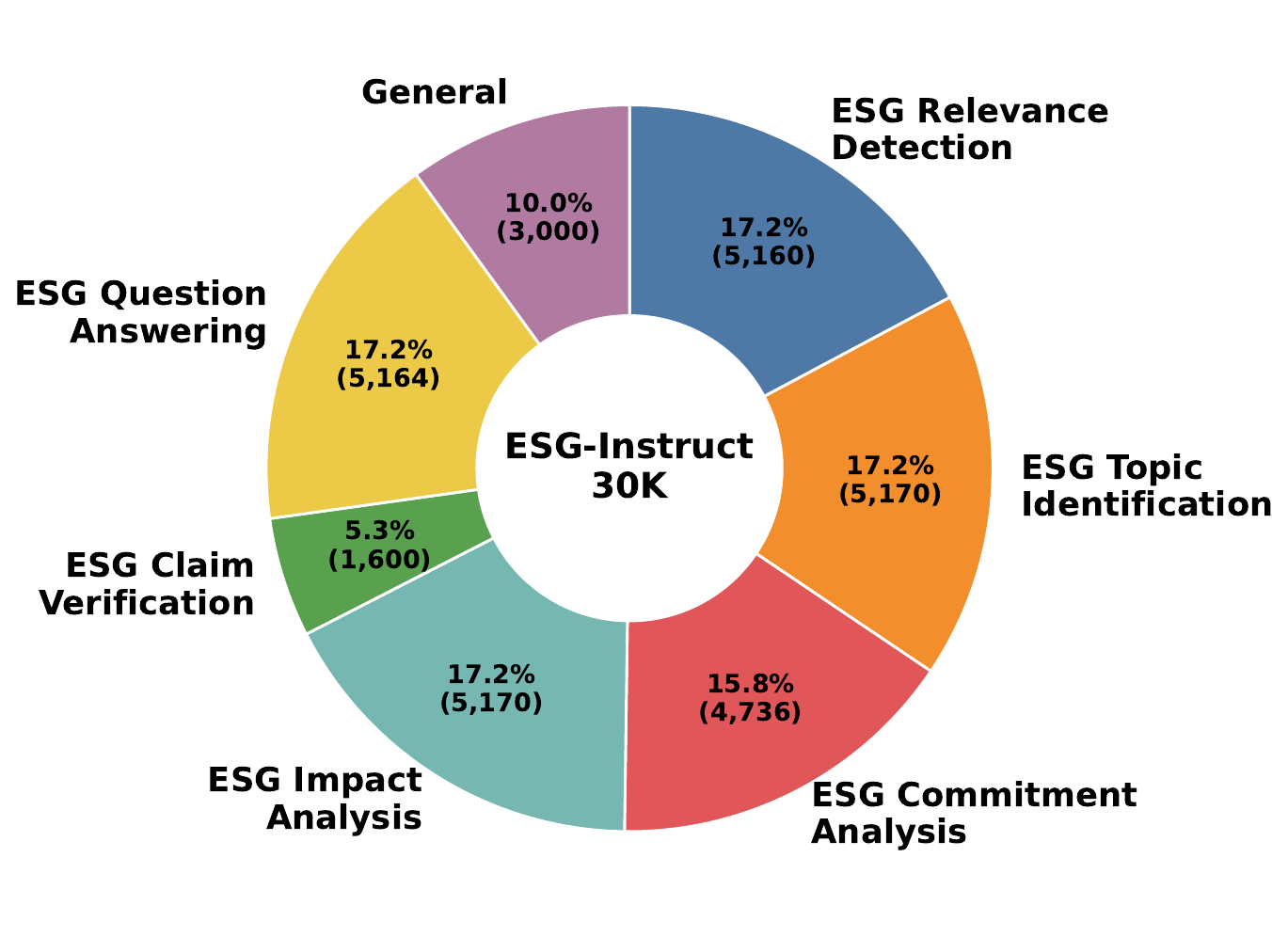}
    \caption{Category distribution of \textsf{ESG-Instruct} 30K.}
    \label{fig:Dist_ESG_instruct}
\end{figure}
\subsection{\textsf{ESG-Instruct}}
\label{app:esg_instruct_details}
The construction of \textsf{ESG-Instruct} proceeds in three main stages: translation, instruction conversion, and final training-data selection. Since high-quality data publicly available in the ESG domain are limited, we include not only human-annotated data but also multilingual ESG data expanded through translation and a limited amount of data that involved LLM assistance during construction. However, these data are used only for training ESG domain adaptation and are kept separate from \textsf{ESGU-Bench}, which is designed for reliable model evaluation.
\begin{table}[t]
\centering
\resizebox{\linewidth}{!}{%
\begin{tabular}{llrr}
\toprule
\textbf{Dataset} & \textbf{Translation Direction} & \textbf{Samples} & \textbf{Time} \\
\midrule
ESG-Kor & Korean $\rightarrow$ English & 118,946 & 130:17:04 \\
Dynamic ESG & Chinese $\rightarrow$ English & 2,200 & 6:24:04 \\
ML-ESG1 & French $\rightarrow$ English & 1,500 & 4:27:33 \\
ML-ESG2 & French $\rightarrow$ English & 1,018 & 3:04:08 \\
ML-ESG2 & Japanese $\rightarrow$ English & 1,121 & 3:01:23 \\
ML-ESG3 & French $\rightarrow$ English & 807 & 2:24:45 \\
ML-ESG3 & Japanese $\rightarrow$ English & 1,553 & 4:14:13 \\
ML-ESG3 & Korean $\rightarrow$ English & 1,000 & 2:53:58 \\
PromiseEval & French $\rightarrow$ English & 800 & 1:54:26 \\
PromiseEval & Japanese $\rightarrow$ English & 800 & 1:56:57 \\
\midrule
\multicolumn{2}{l}{\textbf{Total}} & \textbf{129,745} & \textbf{160:38:31} \\
\bottomrule
\end{tabular}%
}
\caption{Translation time for constructing \textsf{ESG-Instruct} from multilingual ESG datasets.}
\label{tab:translation_time}
\end{table}
\begin{table*}[t]
\centering
\small
\resizebox{\textwidth}{!}{%
\begin{tabular}{p{0.19\linewidth}p{0.68\linewidth}r}
\toprule
Check & Description & Overlap \\
\midrule
Exact input & Exact match over normalized inputs. & 0 \\
Input--target pair & Exact match over normalized input--target pairs. & 0 \\
Near-duplicate (3-gram) & 3-gram Jaccard similarity $\geq 0.80$ within each ESG category. & 0 \\
Near-duplicate (5-gram) & 5-gram Jaccard similarity $\geq 0.80$ within each ESG category. & 0 \\
Source/document & Exact match over available source/document identifiers after filtering generic identifiers. & 0 \\
\bottomrule
\end{tabular}
}
\caption{Contamination checks between \textsf{ESG-Instruct} and \textsf{ESGU-Bench}.}
\label{tab:contamination_check}
\end{table*}
First, non-English ESG datasets are translated so that they can be used for English-based training. We use the \texttt{Gemini 3.1 Flash-Lite} for translation and design a dedicated translation prompt to preserve numbers, named entities, temporal expressions, commitment expressions, and verification-related wording that are important in ESG, sustainability, finance, and corporate reporting contexts. In particular, the prompt instructs the model not to summarize or expand the original meaning and to preserve ambiguity when it exists in the source text. In total, we translate 129,745 non-English ESG samples. The overall processing time is 160 hours, 38 minutes, and 31 seconds, with a total translation cost of approximately \$187. The data scale and processing time for translation are reported in Table~\ref{tab:translation_time}, and an example of the translation prompt is provided below.



\begin{center}
\begin{tcolorbox}[
    breakable,
    enhanced jigsaw,
    width=0.95\linewidth,
    colback=white,
    colframe=black,
    coltitle=white,
    colbacktitle=black,
    title=\textbf{Translation Prompt},
    fonttitle=\bfseries,
    boxrule=0.8pt,
    arc=2mm,
    left=6pt,
    right=6pt,
    top=6pt,
    bottom=6pt,
    before skip=6pt,
    after skip=6pt
]
\small
\textbf{Instruction:}\\
You are a professional translator for ESG, sustainability, finance, and business reports.

\vspace{0.5em}
Translate the following French ESG report text into fluent English.

\vspace{0.5em}
\textbf{Rules:}
\begin{itemize}
    \setlength{\itemsep}{0pt}
    \setlength{\leftmargin}{1.2em}
    \item Preserve meaning exactly. Do not summarize, omit, or add information.
    \item Preserve reporting nuance, commitments, verification wording, dates, figures, percentages, and proper nouns.
    \item Keep ambiguity if present.
    \item Keep paragraph breaks if they exist.
    \item Output only the translation.
    \item Do not add notes, comments, explanations, or quotation marks.
\end{itemize}

\textbf{Source Text:}\\
\texttt{\{text\}}

\vspace{0.5em}
\textbf{Output:}\\
English translation
\end{tcolorbox}
\end{center}
Next, each dataset is converted into a chat-prompt-based instruction format. Each instance consists of a system prompt, task-specific instruction, an input text or evidence context, and a target answer. The target answer preserves the original gold label or gold answer so that the label space and answer structure defined in each dataset remain unchanged. Prompt examples used for each task are provided in Appendix~\ref{app:task_taxonomy_prompt}.

\begin{figure*}[t]
    \centering
    \includegraphics[width=1\linewidth]{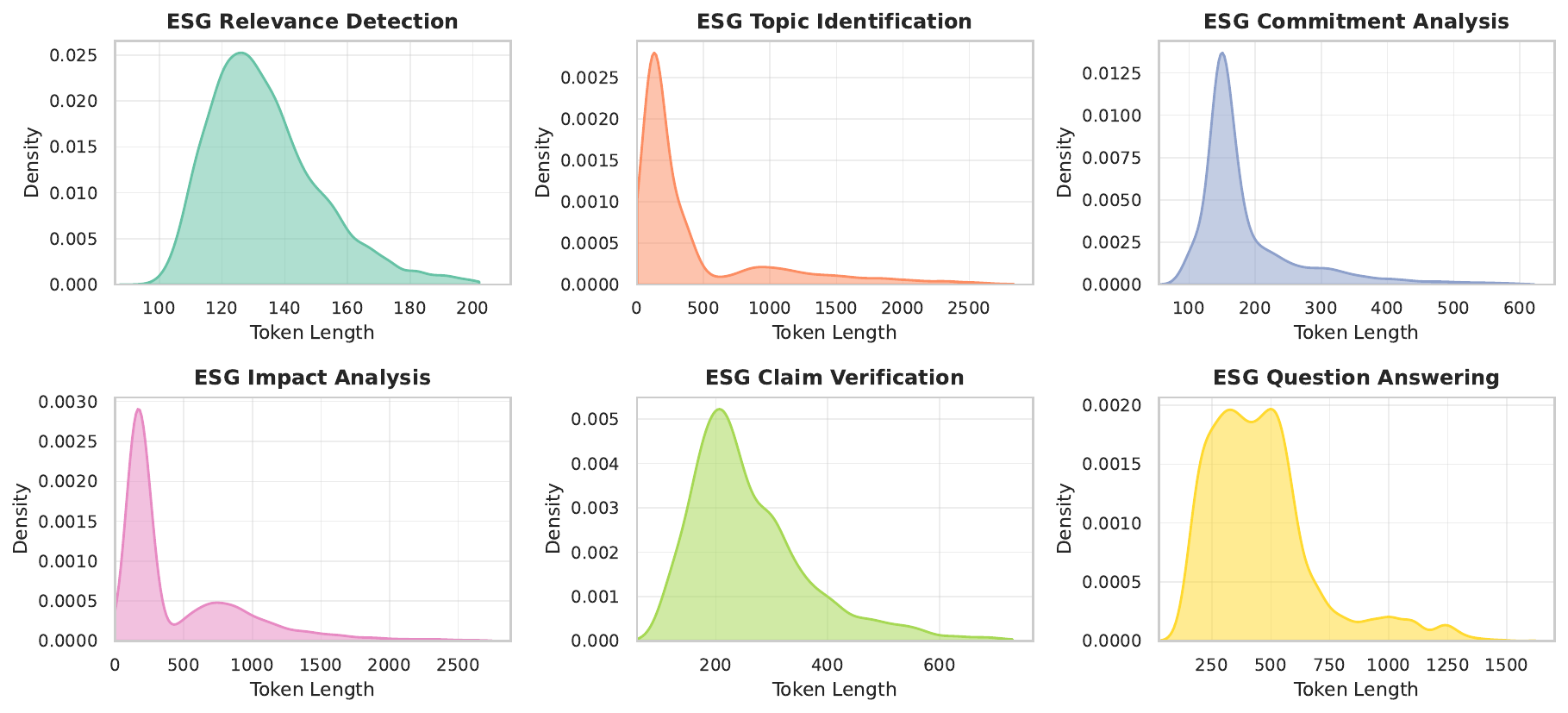}
    \caption{Token length distributions of the \textsf{ESG-Instruct}.}
    \label{fig:Token_Dist_ESG_instruct}
\end{figure*}

Finally, the training data are selected by considering the distributions across the six high-level ESG understanding categories and individual tasks, so that no particular dataset or task is overrepresented. Within each task, we review the label distribution, the alignment between the input and target answer, and the presence of formatting errors. Samples with unclear labels or those that do not match the task objective are excluded. Through this process, we construct the final 30K-scale \textsf{ESG-Instruct}, consisting of 35 ESG instruction tasks and general instruction data~\cite{taori2023alpaca}. The task-level composition of \textsf{ESG-Instruct} and the final number of selected instances are presented in Table~\ref{tab:esg_instruct_composition}. Figure~\ref{fig:Dist_ESG_instruct} shows the category-level distribution of the final 30K training data, and Figure~\ref{fig:Token_Dist_ESG_instruct} shows the token-level distribution of the training instances.

\begin{figure*}[t]
    \centering
    \includegraphics[width=1\linewidth]{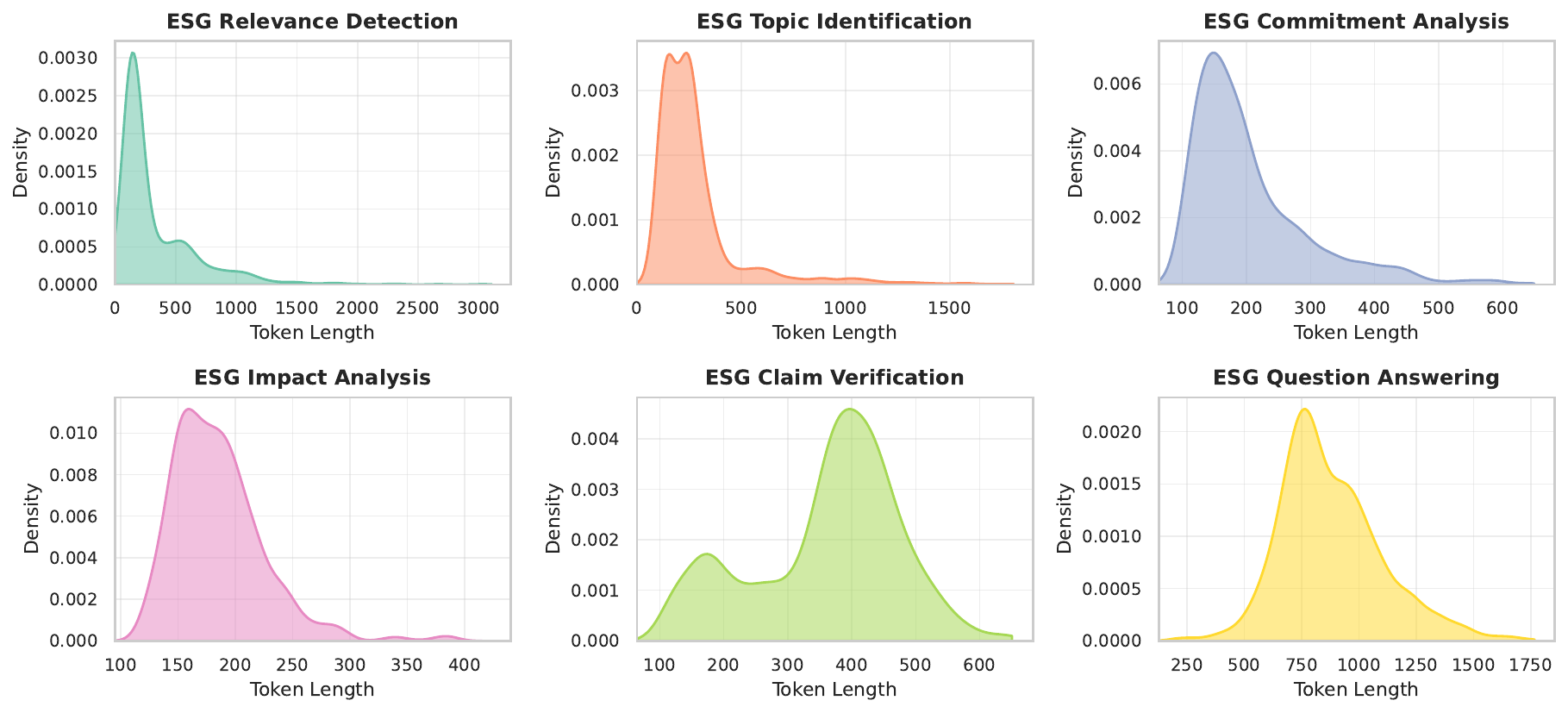}
    \caption{Token length distributions of the \textsf{ESGU-Bench}.}
    \label{fig:Token_Dist_ESGU_bench}
\end{figure*}
\subsection{\textsf{ESGU-Bench}}
\label{app:esgu_details}
\textsf{ESGU-Bench} is constructed with priority on the reliability and comparability of evaluation data. To this end, the benchmark primarily includes datasets with human annotations, expert review, or clearly provided gold answers. In contrast, datasets whose construction mainly involved LLMs or whose label verification process could not be sufficiently confirmed are excluded from benchmark evaluation.

To ensure reproducibility and fair comparison, we use the official test split provided by the original dataset as benchmark instances whenever available. When no official test split is provided, we either use the subset released for evaluation in the original dataset or construct an evaluation subset through data quality review. Most benchmark tasks are converted into a chat-prompt format for model evaluation and are designed in a controlled manner for each task to improve consistency across evaluation conditions. For ESG Question Answering, we retain the original data format to preserve the given question, context, and answer structure~\cite{he2025esgenius}. The task-level composition of \textsf{ESGU-Bench} and the final number of evaluation instances are presented in Table~\ref{tab:esg_benchmark_composition}. Figure~\ref{fig:Token_Dist_ESGU_bench} shows the category-wise token length distribution of benchmark instances.

\subsection{Data Leakage Control}
\label{app:data_leakage}
We carefully separated \textsf{ESG-Instruct} and \textsf{ESGU-Bench} during data construction to prevent train--evaluation contamination. 
Although some dataset families, such as PromiseEval and the ML-ESG series, appear in both resources, this corresponds to dataset-family overlap rather than sample-level leakage. 
In particular, for multilingual datasets, non-English subsets were translated and used for \textsf{ESG-Instruct}, whereas English gold/test subsets were used for \textsf{ESGU-Bench}. 
Thus, belonging to the same dataset family does not imply that the same sample pool is shared between training and evaluation.

Because publicly available high-quality ESG training data are limited, \textsf{ESG-Instruct} includes translated data and limited LLM-assisted data to cover diverse ESG tasks, languages, and source types. 
In contrast, \textsf{ESGU-Bench} was constructed separately for reliable evaluation, focusing on instances with human annotations, expert review, or clear gold answers. 
Whenever available, we used the official test split of the original dataset.

To further verify the independence between the two resources, we conducted contamination checks between the final \textsf{ESG-Instruct} training instances and \textsf{ESGU-Bench} evaluation instances~(Table~\ref{tab:contamination_check}). 
Since non-English training instances were translated into English before training, potential cross-lingual overlaps were also checked in the same English text space through exact and n-gram near-duplicate matching.

\section{ESG Task Taxonomy and Prompt Templates}
\label{app:task_taxonomy_prompt}
We provide the ESG task taxonomy and prompt examples used in this study. While preserving the task definitions and answer structures of diverse ESG datasets, we organize related tasks into six high-level ESG task categories for training-data construction and benchmark evaluation. Since each task differs in input structure, label space, answer format, and output requirements, we design a task-specific prompt template for each task. However, presenting prompts for all tasks would substantially increase the paper length; therefore, we provide representative prompt examples for each of the six high-level ESG task categories.
\subsection{ESG Relevance Detection}
ESG relevance detection is the task of determining whether a given sentence- or document-level text is related to ESG issues. This category includes ESG relevance classification, climate-related text detection, and environmental claim detection. The model must determine whether the input text is substantively connected to ESG-related topics such as environmental, social, governance, climate, sustainability, and corporate responsibility issues.

\begin{promptbox}{ESG Relevance Classification}
\small
\textbf{[System]}\\
You are an ESG sentence classification assistant. Use only the provided English sentence translated from a Korean ESG disclosure. Return valid JSON.

\vspace{0.6em}
\textbf{[User]}\\
Task: Decide whether this corporate disclosure sentence discusses an ESG matter.

\vspace{0.3em}
Valid labels: ESG-related, Not ESG-related.

\vspace{0.3em}
Return JSON with this schema:\newline \texttt{\{"label": string\}}.

\vspace{0.3em}
Sentence: \texttt{\{sentence\}}

\vspace{0.6em}
\textbf{[Assistant]}\\
\texttt{\{"label": "\{label\}"\}}
\end{promptbox}

\subsection{ESG Topic Identification}
ESG topic identification is the task of classifying which ESG area or fine-grained issue an ESG-related text belongs to. This category includes ESG primary category classification, ESG issue classification, SASB issue topic classification, and TCFD recommendation classification. The model must identify the main topic of the input text and select the most appropriate label from predefined ESG categories or topic labels.

\begin{promptbox}{ESG Topic Identification}
\small
\textbf{[System]}\\
You are an ESG sentence classification assistant. Use only the provided English sentence translated from a Korean ESG disclosure. Return valid JSON.

\vspace{0.6em}
\textbf{[User]}\\
Task: Assign the most appropriate ESG pillar to the sentence.

\vspace{0.3em}
Allowed categories: Environmental, Social, Governance, Irrelevant.

\vspace{0.3em}
Return JSON with this schema:\newline \texttt{\{"category": string\}}.

\vspace{0.3em}
Sentence: \texttt{\{sentence\}}

\vspace{0.6em}
\textbf{[Assistant]}\\
\texttt{\{"category": "\{category\}"\}}
\end{promptbox}

\subsection{ESG Commitment Analysis}
ESG commitment analysis is the task of analyzing corporate ESG-related commitments, goals, action plans, reduction targets, and verifiability. This category includes climate commitment/action classification, net-zero reduction target classification, promise status classification, and promise verification timeline classification. The model must distinguish general ESG mentions from actual commitment expressions and assess their specificity, implementation status, target timeline, and verifiability.

\begin{promptbox}{ESG Commitment Analysis}
\small
\textbf{[System]}\\
You are an ESG promise timeline classification assistant. Use only the provided English ESG report passage. Return valid JSON.

\vspace{0.6em}
\textbf{[User]}\\
Task: Classify the verification timeline for the ESG promise in the passage.

\vspace{0.3em}
Valid labels: Already, Less than 2 years, 2 to 5 years, More than 5 years, N/A.

\vspace{0.3em}
Use N/A when the passage does not contain an ESG promise, goal, target, commitment, or verifiable action.

\vspace{0.3em}
Return JSON with this schema:\\ \texttt{\{"verification\_timeline": string\}}.

\vspace{0.3em}
Passage: \texttt{\{passage\}}

\vspace{0.6em}
\textbf{[Assistant]}\\
\texttt{\{"verification\_timeline": "\{verification\_timeline\}"\}}
\end{promptbox}

\subsection{ESG Impact Analysis}
ESG impact analysis is the task of determining the direction, type, level, and duration of the impact that ESG-related events, activities, or statements have on companies, society, or the environment. This category includes climate sentiment classification, impact type classification, impact duration classification, and impact level classification. The model must identify the nature of ESG impact in the input text and analyze whether the impact is positive or negative, short-term or long-term, and how substantial it is.
\begin{promptbox}{ESG Impact Analysis}
\small
\textbf{[System]}\\
You are an ESG impact type classification assistant. Use only the provided English text. Return valid JSON.

\vspace{0.6em}
\textbf{[User]}\\
Task: Determine whether the text describes an ESG opportunity, risk, or cannot be distinguished.

\vspace{0.3em}
Allowed labels: Opportunity, Risk, Cannot distinguish.

\vspace{0.3em}
Return JSON with this schema:\newline \texttt{\{"impact\_type": string\}}.

\vspace{0.3em}
Title: \texttt{\{title\}}

\vspace{0.3em}
Text: \texttt{\{text\}}

\vspace{0.6em}
\textbf{[Assistant]}\\
\texttt{\{"impact\_type": "\{impact\_type\}"\}}
\end{promptbox}

\subsection{ESG Claim Verification}
ESG claim verification is the task of assessing the relationship between an ESG-related claim and its supporting evidence. This category includes climate claim verification, evidence quality classification, and evidence status classification. The model must evaluate whether a given claim is supported, refuted, difficult to determine, or sufficiently grounded by the evidence.
\begin{promptbox}{ESG Claim Verification}
\small
\textbf{[System]}\\
You are a climate claim verification assistant. Use only the provided claim and evidence. Do not use outside knowledge. Return valid JSON.

\vspace{0.6em}
\textbf{[User]}\\
Task: Determine the final verdict for the climate claim using only the provided evidence.

\vspace{0.3em}
Valid labels: SUPPORTS, REFUTES,\newline NOT\_ENOUGH\_INFO, DISPUTED.

\vspace{0.3em}
Return JSON with this schema:\newline \texttt{\{"label": string\}}.

\vspace{0.3em}
Claim: \texttt{\{claim\}}

\vspace{0.3em}
Evidence: \texttt{\{evidence\}}

\vspace{0.6em}
\textbf{[Assistant]}\\
\texttt{\{"label": "\{label\}"\}}
\end{promptbox}

\subsection{ESG Question Answering}
ESG question answering is the task of generating answers to ESG-related questions based on a given context, report, table, or evidence document. This category includes factoid QA, non-factoid QA, and table QA based on ESG reports and sustainability documents. The model must understand the intent of the question and generate an answer grounded in the provided context or evidence.

\begin{promptbox}{ESG Question Answering}
\small
\textbf{[System]}\\
You are a sustainability report question-answering assistant. Use only the provided context passage to answer descriptively. Return valid JSON.

\vspace{0.6em}
\textbf{[User]}\\
Task: Answer the ESG-related question based strictly on the provided context.

\vspace{0.3em}
Return JSON with this schema:\newline \texttt{\{"answer": string\}}.

\vspace{0.3em}
Context: \texttt{\{context\}}

\vspace{0.3em}
Question: \texttt{\{question\}}

\vspace{0.6em}
\textbf{[Assistant]}\\
\texttt{\{"answer": "\{answer\}"\}}
\end{promptbox}

\section{Implementation Details}
\label{app:implementation}
We compare models from three groups: large proprietary LLMs, open-source LLMs, and specialized ESG-oriented models. 
The proprietary LLM group includes recent API-based GPT and DeepSeek models. 
The open-source LLM group includes models from diverse families and sizes, such as Qwen, Llama, Gemma, Mistral, and DeepSeek-distilled models. 
The specialized model group includes SusGen-GPT. 
For \textsf{ESGPT}, we used Qwen3-8B as the base model and fine-tuned it using LoRA-based PEFT adapter training~\cite{hu2022lora}. 
During training, we used a maximum sequence length of 4096 tokens, 3 epochs, a per-device batch size of 1, gradient accumulation steps of 16, and an effective batch size of 16. 
The learning rate was set to $1\times10^{-4}$ with a cosine scheduler, a warmup ratio of 0.03, and no weight decay. 
Training was performed in bf16 precision with the AdamW optimizer. 
For LoRA, we used rank 64, alpha 128, dropout 0.05, and applied adapters to \texttt{q\_proj}, \texttt{k\_proj}, \texttt{v\_proj}, \texttt{o\_proj}, \texttt{gate\_proj}, \texttt{up\_proj}, and \texttt{down\_proj}. 
During inference, local models were loaded with 4-bit quantization by default and evaluated with a batch size of 4, a maximum input length of 4096 tokens, and maximum generation lengths of 256 tokens for classification tasks and 1024 tokens for QA or generation tasks. We used temperature 0.2 and top-$p$ 0.90 for decoding. API-based models were evaluated under the same temperature setting. All models were evaluated under the same evaluation protocol, with accuracy used as the evaluation metric across tasks. A JSON parser is first used to extract the required answer field from model outputs; if parsing fails, deterministic raw-text extraction is applied as a fallback. The extracted output is matched against the task-specific valid label set, and invalid or unmappable outputs are scored as incorrect. For ESG QA, formulated as multiple-choice QA, exact-match accuracy over predefined answer options is used. All experiments were conducted on a single NVIDIA A100 GPU. 

\section{Details of Experimental Results}
\label{app:additional_results}
This section provides additional detailed experimental results. While the main text focuses on comparisons with representative models on \textsf{ESGU-Bench}, this appendix reports the ranking of all models, category-level performance, and performance changes according to the training data scale in greater detail.

\subsection{\textsf{ESGU-Bench} Results}
\label{app:full_results}
\begin{figure*}[t]
    \centering
    \includegraphics[width=1\linewidth]{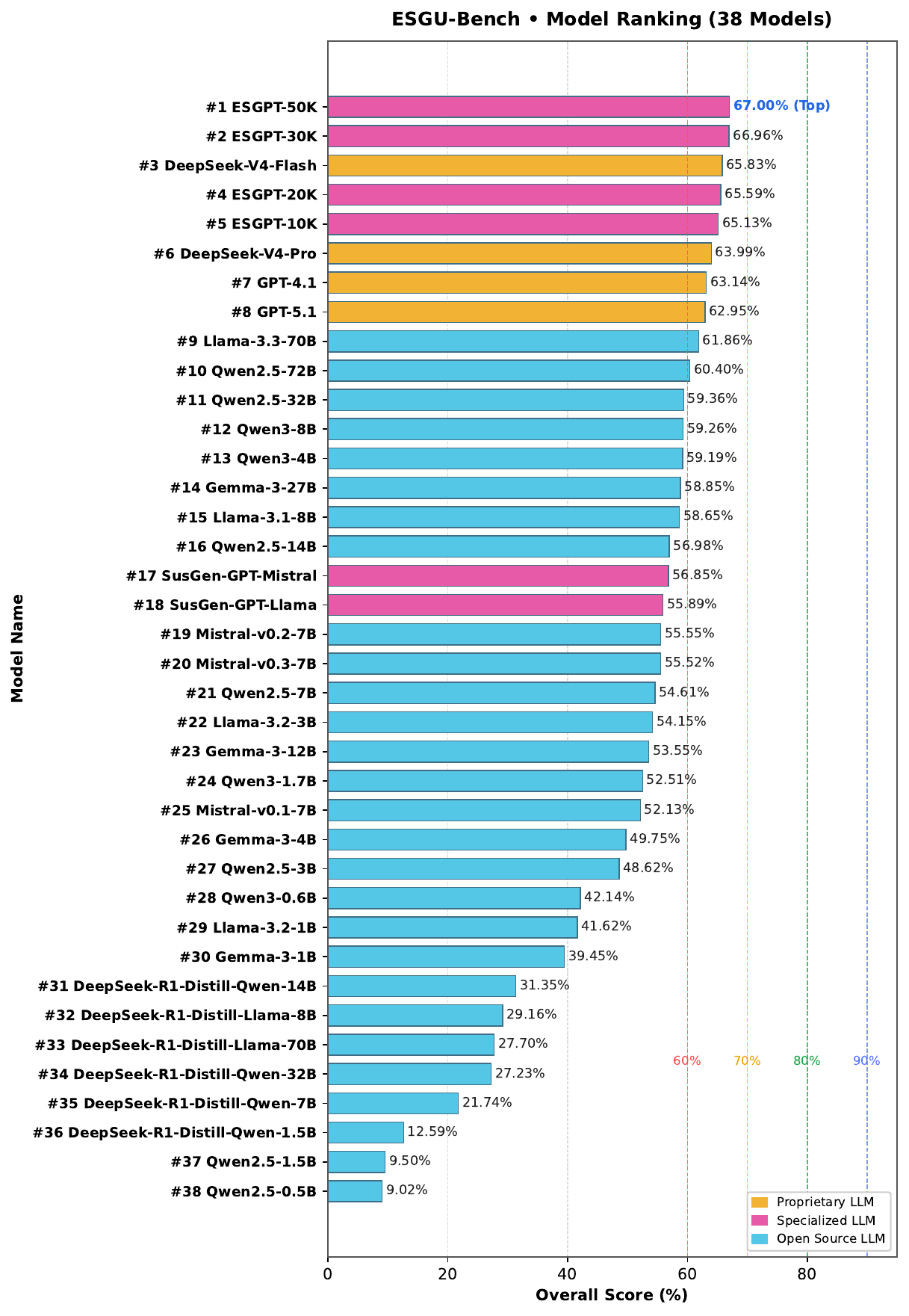}
    \caption{Overall model ranking on \textsf{ESGU-Bench} across 38 evaluated models. Bars indicate overall scores, and colors denote model groups: proprietary LLMs, specialized LLMs, and open-source LLMs.}
    \label{fig:modelrank}
\end{figure*}

Figure~\ref{fig:modelrank} shows the ranking of all 38 evaluated models on \textsf{ESGU-Bench} based on the overall score. The 38 models include the 35 LLMs discussed in the main text as well as additional \textsf{ESGPT} variants trained with different data scales, such as 10K, 20K and 50K. All \textsf{ESGPT} data-scale variants were trained with the same fixed random seed and training configuration to ensure a controlled comparison across data scales. In the overall ranking, the \textsf{ESGPT} models are placed among the top-performing models. Despite their 8B scale, they achieve higher overall scores than large proprietary LLMs, including DeepSeek-V4-Flash, DeepSeek-V4-Pro, GPT-4.1, and GPT-5.1. This suggests that, for ESG understanding tasks, instruction data designed for the ESG domain and task-specific adaptation can serve as important performance factors, beyond model size or general capability.

Table~\ref{tab:overall_performance} reports the full category-level performance across the six ESG understanding categories. For the final \textsf{ESGPT} result, we additionally report the mean $\pm$ standard deviation over five random seeds to assess performance stability. \textsf{ESGPT} achieves the highest overall score and records the best performance in ESG relevance detection, ESG topic identification, ESG commitment analysis, and ESG claim verification. This shows that \textsf{ESGPT} effectively adapts to tasks where understanding ESG disclosure contexts and task-specific label spaces is important, such as identifying ESG-related text, classifying ESG topics, interpreting corporate commitments and targets, and performing evidence-based verification. In contrast, DeepSeek-V4-Flash achieves the best performance in ESG impact analysis, while GPT-5.1 performs best in ESG question answering. This suggests that, for some tasks, the general reasoning ability and broad prior knowledge of large proprietary LLMs may still provide advantages.

Overall, \textsf{ESGPT}, despite being a relatively small 8B model, achieves performance that is competitive with or superior to that of large LLMs. This demonstrates that ESG-specialized models can be both efficient and effective. However, the overall results also show that both large proprietary LLMs and open-source LLMs still have limited performance in certain ESG categories. In particular, tasks that require complex contextual understanding and evidence-based judgment, such as ESG question answering, impact interpretation, and claim verification, leave substantial room for improvement. Therefore, \textsf{ESGU-Bench} can be viewed as a challenging benchmark that reveals both the limitations of current LLMs in ESG understanding tasks and directions for future improvement.




\subsection{Data Scale Analysis}
\label{app:scale_ablation}
Table~\ref{tab:scale_ablation} compares the performance of \textsf{ESGPT} under different training data scales using Qwen3-8B and Llama-3.1-8B-Instruct as base models. For both base models, the overall score generally improves as the training data scale increases from 10K to 30K and 50K. The Qwen3-8B-based model achieves the highest overall score of 67.00\% in the 50K setting, while the 30K setting shows a comparable score of 66.96\%. The Llama-3.1-8B-Instruct-based model also achieves its best performance in the 50K setting with 64.72\%, followed by the 30K setting with 64.24\%. In contrast, the full-data setting leads to a substantial drop in the overall score for both base models.

At the category level, the 30K or 50K settings show strong performance on most tasks, whereas ESG claim verification achieves the best performance in the 20K setting for both base models. This can be attributed to the relatively limited amount of training data available for ESG claim verification, which makes it difficult to maintain the balance of this task as the overall data scale increases. These results show that, rather than simply using more data, data construction and training strategies that consider task balance and label quality are critical for ESG instruction tuning.


\section{Qualitative Case Studies}
\label{app:case_studies}
This section presents five qualitative case studies observed in \textsf{ESGU-Bench}. Each case study includes the prompt, input text, gold label, and model responses used in the actual evaluation. Case Studies 1--3 are examples where \textsf{ESGPT} predicts the correct answer while large API-based LLMs fail, whereas Case Studies 4--5 are challenging cases where all comparison models fail. Through these cases, we analyze the strengths of ESG-specialized LLMs in interpreting ESG contexts, disclosure expressions, and task-specific label boundaries, as well as the types of ESG understanding that remain challenging for current models.


\begin{qualcasebox}{Case 1. Commitment Signal}
\small
\textbf{[User]}

\footnotesize
Classify whether this report excerpt includes an ESG target, commitment, or promise.\\
Allowed labels: Yes, No.

\vspace{0.6em}
\textbf{[Input]}

\footnotesize
Qualification, Orientation and Training All safety sensitive positions and supervisory staff receive extensive safety training, and all operations personnel are enrolled in a competency program that ensures only those deemed competent through training, exams, observation and sign-off by designated assessors can perform safety sensitive tasks unsupervised. Specific competency profiles are developed for each role.

\tcblower

\small
\textbf{[Gold and Model Answers]}

\vspace{0.2em}
\footnotesize
\begin{tabularx}{1.0\linewidth}{@{}Y C{0.25\linewidth} C{0.25\linewidth}@{}}
\hline
\textbf{Model} & \textbf{Answer} & \textbf{Result} \\
\hline
Gold & \goldans{Yes} & -- \\
\hline
DeepSeek-V4-Flash & \wrongans{No} & \xmarkred \\
DeepSeek-V4-Pro & \wrongans{No} & \xmarkred \\
GPT-4.1 & \wrongans{No} & \xmarkred \\
GPT-5.1 & \wrongans{No} & \xmarkred \\
\textsf{ESGPT} & \correctans{Yes} & \cmarkblue \\
\hline
\end{tabularx}
\end{qualcasebox}
Case 1 shows that ESG commitments are not always expressed through explicit goals, numerical targets, or declarative phrases such as ``we commit to.'' The input paragraph describes training, competency verification, testing, observation, and approval procedures for safety-sensitive roles, which can be viewed as ESG-related action signals indicating that the company operates a safety management system in practice. Large API-based LLMs answer No by focusing on the absence of explicit commitment expressions, whereas \textsf{ESGPT} answers Yes by capturing procedural and operational commitment signals that frequently appear in ESG reports. This demonstrates that an ESG-specialized LLM can more appropriately interpret implicit action signals in corporate disclosure texts.

\begin{qualcasebox}{Case 2. Impact Severity}
\small
\textbf{[User]}

\footnotesize
Analyze the provided ESG news text and choose its impact severity.\\
Allowed labels: Low, Medium, High.

\vspace{0.6em}
\textbf{[Input]}

\footnotesize
\textbf{Title:} A.P. Møller Acquires Sustainable Food Packaging Company Faerch Group

\vspace{0.3em}
\textbf{Article:} Faerch is a leading European manufacturer of sustainable food packaging solutions within the Ready Meals, Protein, Food-to-Go and Dairy segments, providing circular solutions for rigid food packaging, with products made from up to 100\% recycled content and fully recyclable to new food grade products. According to A.P. Moller, Faerch is also the only food packaging company in the world to operate as an integrated recycler.

\tcblower

\small
\textbf{[Gold and Model Answers]}

\vspace{0.2em}
\footnotesize
\begin{tabularx}{1.0\linewidth}{@{}Y C{0.25\linewidth} C{0.25\linewidth}@{}}
\hline
\textbf{Model} & \textbf{Answer} & \textbf{Result} \\
\hline
Gold & \goldans{High} & -- \\
\hline
DeepSeek-V4-Flash & \wrongans{Medium} & \xmarkred \\
DeepSeek-V4-Pro & \wrongans{Medium} & \xmarkred \\
GPT-4.1 & \wrongans{Low} & \xmarkred \\
GPT-5.1 & \wrongans{Low} & \xmarkred \\
\textsf{ESGPT} & \correctans{High} & \cmarkblue \\
\hline
\end{tabularx}
\end{qualcasebox}
Case 2 shows that ESG impact analysis is not merely a task of determining ESG relevance, but also requires assessing the ESG significance and impact level of an event. The input article describes the acquisition of a sustainable food packaging company, which provides circular packaging solutions, products made from up to 100\% recycled content, full recyclability into food-grade packaging, and an integrated recycling system. API-based LLMs interpret this as a general ESG-related corporate acquisition and assign lower ratings of \texttt{Medium} or \texttt{Low}. In contrast, \textsf{ESGPT} predicts \texttt{High} by reflecting the strategic ESG significance of an event that combines circular economy practices with a recycling-based business transition. This suggests that ESG-specialized models may be better suited to assessing impact severity beyond the surface topic of an event.

\begin{qualcasebox}{Case 3. ESG Reasoning Boundary}
\small
\textbf{[User]}

\footnotesize
Answer the multiple-choice ESG reasoning question with a single option letter.\\
Valid options: A, B, C, D, Z.

\vspace{0.6em}
\textbf{[Input]}

\footnotesize
\textbf{Question:} What is the primary reason the ISSB rejected the hierarchical structure approach despite its potential to enhance comparability of digital information across sustainability-related disclosure standards, as described in the IFRS Taxonomy: IFRS Sustainability Disclosure Taxonomy 2024 -- IFRS S1 General Requirements for Disclosure of Sustainability-related Financial Information and IFRS S2 Climate-related Disclosures?

\vspace{0.3em}
\textbf{Options:}\\
A: The hierarchical structure would lead to fragmented tagged information that lacks sufficient context at lower levels.\\
B: Tagging systems are too new and untested to handle the complexity introduced by a hierarchical structure.\\
C: Regulators might require entities to use varying levels of elements, resulting in inconsistent digital information.\\
D: Elements at paragraph level were deemed insufficiently detailed, while lower-level elements lacked standalone utility.\\
Z: Not sure

\tcblower

\small
\textbf{[Gold and Model Answers]}

\vspace{0.2em}
\footnotesize
\begin{tabularx}{1.0\linewidth}{@{}Y C{0.25\linewidth} C{0.25\linewidth}@{}}
\hline
\textbf{Model} & \textbf{Answer} & \textbf{Result} \\
\hline
Gold & \goldans{C} & -- \\
\hline
DeepSeek-V4-Flash & \wrongans{A} & \xmarkred \\
DeepSeek-V4-Pro & \wrongans{A} & \xmarkred \\
GPT-4.1 & \wrongans{A} & \xmarkred \\
GPT-5.1 & \wrongans{A} & \xmarkred \\
\textsf{ESGPT} & \correctans{C} & \cmarkblue \\
\hline
\end{tabularx}
\end{qualcasebox}
Case 3 shows that, in ESG question answering, it is important to distinguish between explanations that are generally plausible and those that align with the correct answer criterion. Option A, selected by the API models, is a plausible explanation in that hierarchical tagging may cause context loss at lower-level tags. However, the correct answer required by this question is not about the general limitation of hierarchical structures, but about the reduced consistency of digital information when regulators require the use of taxonomy elements at different levels. \textsf{ESGPT} predicts C, which aligns with the benchmark's answer criterion, rather than the generally plausible answer selected by the API models. This shows that an ESG-specialized LLM can better distinguish subtle semantic differences among options and task-specific decision boundaries in ESG question answering tasks.


\begin{qualcasebox}{Case 4. Commitment Boundary}
\small
\textbf{[User]}

\footnotesize
Review the paragraph and return whether it concerns climate commitments and actions.\\
Valid labels: yes, no.

\vspace{0.6em}
\textbf{[Input]}

\footnotesize
Furthermore, according to the United Nations Environment Programme's (UNEP) latest Emissions Gap Report, 1.5$^\circ$C will involve ``expanding renewable energy for electrification, phasing out coal for rapid decarbonisation of the energy system, decarbonising transport with a focus on electric mobility, decarbonising energy-intensive industry and avoiding future emissions, while improving energy access''; all very challenging actions.

\tcblower

\small
\textbf{[Gold and Model Answers]}

\vspace{0.2em}
\footnotesize
\begin{tabularx}{1.0\linewidth}{@{}Y C{0.25\linewidth} C{0.25\linewidth}@{}}
\hline
\textbf{Model} & \textbf{Answer} & \textbf{Result} \\
\hline
Gold & \goldans{no} & -- \\
\hline
DeepSeek-V4-Flash & \wrongans{yes} & \xmarkred \\
DeepSeek-V4-Pro & \wrongans{yes} & \xmarkred \\
GPT-4.1 & \wrongans{yes} & \xmarkred \\
GPT-5.1 & \wrongans{yes} & \xmarkred \\
\textsf{ESGPT} & \wrongans{yes} & \xmarkred \\
\hline
\end{tabularx}
\end{qualcasebox}
Case 4 shows a case where all models overreact to climate-action expressions. The input paragraph contains strong climate-action terms such as renewable energy, phasing out coal, decarbonising transport, and electric mobility. However, the sentence does not describe a commitment by a specific company or reporting entity, but rather a general climate transition pathway presented in a UNEP report. In other words, it explains the need for climate action, but does not indicate an explicit commitment or action by the reporting entity. This case shows that distinguishing salient climate keywords from the actual commitment boundary remains challenging for all models, including \textsf{ESGPT}.


\begin{qualcasebox}{Case 5. Sentiment Boundary}
\small
\textbf{[User]}

\footnotesize
Classify the climate sentiment of the given paragraph.\\  
Allowed labels: risk, neutral, opportunity.

\vspace{0.6em}
\textbf{[Input]}

\footnotesize
The important initiatives taken by DNP Group in moving toward a decarbonized society include reducing the amount of CO2-generating energy used, namely energy conservation, switching to low CO2-emission fuels, and introducing renewable energy sources.

\tcblower

\small
\textbf{[Gold and Model Answers]}

\vspace{0.2em}
\footnotesize
\begin{tabularx}{1.0\linewidth}{@{}Y C{0.25\linewidth} C{0.25\linewidth}@{}}
\hline
\textbf{Model} & \textbf{Answer} & \textbf{Result} \\
\hline
Gold & \goldans{neutral} & -- \\
\hline
DeepSeek-V4-Flash & \wrongans{opportunity} & \xmarkred \\
DeepSeek-V4-Pro & \wrongans{opportunity} & \xmarkred \\
GPT-4.1 & \wrongans{opportunity} & \xmarkred \\
GPT-5.1 & \wrongans{opportunity} & \xmarkred \\
\textsf{ESGPT} & \wrongans{opportunity} & \xmarkred \\
\hline
\end{tabularx}
\end{qualcasebox}
Case 5 shows a case where all models interpret positive climate-action expressions as \texttt{opportunity}. The input paragraph describes how DNP Group implemented energy savings, a shift to low-carbon fuels, and the adoption of renewable energy to move toward a decarbonized society. However, the paragraph does not explicitly describe business opportunities such as corporate growth, cost reduction, competitiveness enhancement, or market expansion. Therefore, under the benchmark criterion, the sentence is closer to a neutral description of climate-related actions, and the gold label is \texttt{neutral}. This case shows that distinguishing mentions of positive ESG actions from opportunity framing from a corporate perspective remains challenging.

Overall, Cases 1--3 show that ESG-specialized LLMs can more appropriately interpret implicit signals in ESG contexts, the impact severity of events, and subtle semantic differences among answer options. In contrast, Cases 4--5 show that current models may still overreact to ESG/climate-related keywords or positive expressions, and may struggle to distinguish explicit commitments from general climate-action statements or to identify sentiment label boundaries. These findings confirm that \textsf{ESGU-Bench} provides an evaluation setting that requires fine-grained semantic distinctions in ESG contexts and task-specific judgment, beyond simple ESG keyword detection. At the same time, current models still show limitations in fine-grained ESG judgments, such as explicit commitment detection, sentiment framing, and reasoning boundaries, suggesting the need to improve ESG-specific reasoning and label-boundary awareness.



\section{TCFD Report Generation}
\label{app:report_generation}
We qualitatively compare the TCFD-style reports generated by \textsf{ESGPT} and SusGen-GPT at the report level. A TCFD-style report is constructed by generating answers to 11 questions corresponding to the major disclosure areas of the TCFD recommendations, using evidence retrieved from corporate sustainability/ESG reports, and organizing them into a single report. SusGen-GPT is an LLM specialized for sustainability report generation, trained with both financial NLP tasks and ESG report generation, and is used as a strong domain-specific baseline~\cite{wu2025susgen}.

For analysis, we use NYSE\_PBR\_2016, which corresponds to the 2016 report of Petrobras, a company listed on the New York Stock Exchange~\cite{ni2023chatreport}. This report contains ESG information related to governance, compliance, and risk management, but climate-specific disclosure is unclear for some TCFD items. It is therefore suitable for examining how the two models use general ESG evidence when generating climate-related disclosure. Table~\ref{tab:pbr_side_by_side_report} compares the generation results of the two models on NYSE\_PBR\_2016.

The results show that \textsf{ESGPT} tends to preserve the limitations of the provided evidence when climate-specific disclosure is unclear. For example, in items such as board oversight, management roles, and risk identification processes, \textsf{ESGPT} does not directly infer climate-related oversight or climate-specific risk processes from general governance or risk management information. In contrast, SusGen-GPT generates longer and more report-like responses, but sometimes extends general ESG information into climate-specific disclosure.

This difference is particularly evident in the TCFD 10 item on Scope emissions. \textsf{ESGPT} states that Scope 1 and Scope 2 emissions are disclosed, while distinguishing that Scope 3 emissions are not explicitly mentioned. In contrast, SusGen-GPT first states that the organization discloses Scope 1, Scope 2, and Scope 3 emissions, and then adds within the same answer that explicit Scope 3 data are not provided. This overstates disclosure coverage and introduces an internal inconsistency. Similar tendencies are observed in TCFD 1 and TCFD 9: SusGen-GPT links the Petrobras Corruption Prevention Program to climate-related risk oversight and states that a comprehensive set of metrics exists despite the lack of specific quantitative indicators. By contrast, \textsf{ESGPT} uses more constrained wording when the evidence is insufficient to support a climate-specific claim.

Overall, SusGen-GPT has the advantage of producing report-style responses, but it may over-claim or generate internally inconsistent statements when the evidence is indirect or incomplete. In contrast, \textsf{ESGPT} more carefully preserves ESG disclosure boundaries and produces more evidence-grounded, conservative reports. Since ESG report generation is closely tied to corporate disclosure, these results suggest that faithfulness to evidence is critical for reliable ESG report generation.

\begin{table*}[t]
\centering
\scriptsize
\setlength{\tabcolsep}{4pt}
\renewcommand{\arraystretch}{1.12}
\resizebox{\textwidth}{!}{%
\begin{tabular}{p{0.48\textwidth}|p{0.48\textwidth}}
\hline
\textbf{\textsf{ESGPT}} & \textbf{SusGen-GPT} \\
\hline

\multicolumn{2}{l}{\textbf{TCFD 1. Board Oversight of Climate-related Risks and Opportunities}} \\
\hline
The board oversees climate-related risks and opportunities indirectly through its focus on operational safety and financial leverage reduction. 
\textcolor{caseblue}{However, the sustainability report does not explicitly link these strategies to climate-specific measures or provide details on how the board directly addresses climate risks.}
&
The board oversees climate-related risks and opportunities through strategic goals and governance mechanisms. 
It also monitors the Petrobras Corruption Prevention Program, which \textcolor{casered}{likely includes climate-related risks}. 
\textcolor{casered}{However, the evidence does not explicitly state that climate-related risks are addressed in those meetings or reports.}
\\
\hline

\multicolumn{2}{l}{\textbf{TCFD 2. Management Role in Assessing and Managing Climate-related Risks}} \\
\hline
\textcolor{caseblue}{The evidence does not explicitly link management's role to climate-related risks and opportunities.} 
The disclosure focuses instead on general business strategy, management system outcomes, compliance, governance, prospects, and challenges.
&
Management plays a significant role in assessing and managing climate-related risks and opportunities. 
This is inferred from sections discussing compliance, prospects, challenges, governance, and management system results. 
However, the report does not provide detailed strategies or quantitative evidence for how these risks are addressed.
\\
\hline

\multicolumn{2}{l}{\textbf{TCFD 3. Climate-related Risks and Opportunities over Time Horizons}} \\
\hline
The organisation identifies risks that may jeopardize strategic goals, business execution, regulatory compliance, ethics, market conditions, credit, and liquidity. 
These risks are prioritized based on potential consequences and probability of occurrence.
&
The organization identifies climate-related risks across short-, medium-, and long-term horizons. 
Short-term risks include operational risks, medium-term risks involve compliance, and long-term risks relate to strategic goals. 
\textcolor{casered}{However, explicit climate-related opportunities are not identified in the evidence.}
\\
\hline

\multicolumn{2}{l}{\textbf{TCFD 4. Impact on Strategy, Financial Performance, and Planning}} \\
\hline
Climate-related risks and opportunities are discussed through broader operational challenges, regulatory compliance, and market volatility. 
\textcolor{caseblue}{The evidence is vague and focuses more on general compliance and operational risks than on specific climate-related financial impacts.}
&
\textcolor{casered}{Climate-related risks and opportunities significantly impact} the organization's business strategy, economic and financial performance, and financial planning. 
The answer links this claim to general risks and management-system disclosures, but notes that quantitative financial evidence is not provided.
\\
\hline

\multicolumn{2}{l}{\textbf{TCFD 5. Strategy Resilience under Climate Scenarios}} \\
\hline
The organisation's strategy is aligned with its risk management system and integrated into strategic decisions. 
\textcolor{caseblue}{However, the evidence does not explicitly address resilience under different climate scenarios, including a 2$^\circ$C or lower scenario, or physical climate risks.}
&
The strategy shows integrated risk management and alignment with the Strategic Plan. 
However, the evidence does not provide specific details on climate physical risks or resilience under a 2$^\circ$C scenario.
\\
\hline

\multicolumn{2}{l}{\textbf{TCFD 6. Processes for Identifying and Assessing Climate-related Risks}} \\
\hline
The organisation identifies risks through interviews, online research, and strategic analysis. 
Risks are prioritized based on consequences and probability of occurrence. 
\textcolor{caseblue}{The process for identifying climate-related risks is not explicitly detailed.}
&
The organization uses a multi-step process involving broad diagnosis, online research, face-to-face interviews with executives, Strategic Plan alignment, business-plan risk identification, social-risk analysis, and board monitoring. 
\textcolor{casered}{However, the evidence does not provide specific details on how climate-related risks are identified within the broader risk process.}
\\
\hline

\multicolumn{2}{l}{\textbf{TCFD 7. Management of Climate-related Risks}} \\
\hline
The organization manages climate-related risks by monitoring global alternatives for low-carbon energy, adapting projects and facilities, and disclosing annual atmospheric emissions data from nearly 30,000 sources. 
These efforts were recognized through a high CDP score in 2016.
&
The organization manages climate-related risks through strategic planning, risk responses, monitoring low-carbon energy models, and adapting projects and facilities. 
The answer also notes that the evidence does not provide detailed quantitative strategies for climate-risk mitigation.
\\
\hline

\multicolumn{2}{l}{\textbf{TCFD 8. Integration into Overall Risk Management}} \\
\hline
Climate-related risks are integrated into overall risk management through alignment with the Strategic Plan, risk identification and prioritization, and structured tools such as a social risk matrix. 
These processes support systematic consideration of risks in strategic decisions.
&
Climate-related risks are integrated into overall risk management through Strategic Plan alignment, periodic reporting to the Executive Committee of Risks, social-risk structures, and executive risk oversight. 
\textcolor{casered}{However, the evidence does not explicitly mention climate-related risk integration, only broader risk integration.}
\\
\hline

\multicolumn{2}{l}{\textbf{TCFD 9. Metrics for Assessing Climate-related Risks and Opportunities}} \\
\hline
\textcolor{caseblue}{The organisation does not explicitly disclose specific metrics} for assessing climate-related risks and opportunities. 
Instead, it focuses on broader governance and compliance strategies, which limits clarity on how climate risks are quantified or managed.
&
The organization uses \textcolor{casered}{a comprehensive set of metrics} to assess climate-related risks and opportunities. 
However, \textcolor{casered}{the specific metrics and quantitative values are not explicitly detailed} in the evidence, limiting a thorough assessment.
\\
\hline

\multicolumn{2}{l}{\textbf{TCFD 10. Scope 1, Scope 2, and Scope 3 GHG Emissions}} \\
\hline
The organisation discloses Scope 1 and Scope 2 emissions. 
Scope 1 emissions decreased from 77.4 million metric tons in 2015 to 66.1 million metric tons in 2016, while Scope 2 emissions decreased from 0.7 to 0.4 million metric tons. 
\textcolor{caseblue}{Scope 3 emissions are not explicitly mentioned.}
&
The organization \textcolor{casered}{discloses Scope 1, Scope 2, and Scope 3 greenhouse gas emissions}. 
Scope 1 emissions were 66.1 million tCO2e and Scope 2 emissions were 0.4 million tCO2e in 2016. 
\textcolor{casered}{The report does not provide explicit Scope 3 emissions data.}
\\
\hline

\multicolumn{2}{l}{\textbf{TCFD 11. Targets and Performance against Climate-related Risks and Opportunities}} \\
\hline
\textcolor{caseblue}{The organization does not explicitly mention specific targets} for understanding, quantifying, or benchmarking climate-related risks and opportunities. 
Instead, it focuses on broader sustainability initiatives and governance or compliance strategies.
&
The organization provides governance, business strategy, and management system information. 
However, the evidence does not explicitly state specific climate-related risks and opportunities targets or KPIs, making it unclear how the organization performs against them.
\\
\hline

\end{tabular}%
}
\caption{Side-by-side comparison of TCFD-style report generation outputs from \textsf{ESGPT} and SusGen-GPT for NYSE\_PBR\_2016. The outputs are compressed for readability, with blue highlights indicating evidence-bounded statements and red highlights indicating potential hallucination risks.}
\label{tab:pbr_side_by_side_report}
\end{table*}

\begin{table*}[t]
\centering
\scriptsize
\setlength{\tabcolsep}{3.0pt}
\renewcommand{\arraystretch}{1.08}
\resizebox{\textwidth}{!}{%
\begin{tabular}{l|l|c|cccccc|c}
\hline
\textbf{Type} & \textbf{Model} & \textbf{Size}
& \makecell[c]{\textbf{ESG Relevance}\\\textbf{Detection}}
& \makecell[c]{\textbf{ESG Topic}\\\textbf{Identification}}
& \makecell[c]{\textbf{ESG Commitment}\\\textbf{Analysis}}
& \makecell[c]{\textbf{ESG Impact}\\\textbf{Analysis}}
& \makecell[c]{\textbf{ESG Claim}\\\textbf{Verification}}
& \makecell[c]{\textbf{ESG}\\\textbf{QA}}
& \textbf{Overall} \\
\hline

\multirow{28}{*}{\makecell[c]{Open\\Source\\LLM}}
& Qwen2.5 & 0.5B & 0.00 & 0.04 & 0.06 & 0.00 & 0.17 & 53.87 & 9.02 \\
& Qwen2.5 & 1.5B & 0.19 & 0.13 & 0.06 & 0.00 & 1.80 & 54.84 & 9.50 \\
& Qwen2.5 & 3B & 55.43 & 22.26 & 53.03 & 57.90 & 44.37 & 58.71 & 48.62 \\
& Qwen2.5 & 7B & 64.59 & 35.31 & 55.66 & 59.63 & 49.68 & 62.76 & 54.61 \\
& Qwen2.5 & 14B & 69.78 & 32.93 & 61.32 & 63.33 & 52.55 & 61.97 & 56.98 \\
& Qwen2.5 & 32B & 80.26 & 36.36 & 57.90 & \underline{68.27} & 52.98 & 60.39 & 59.36 \\
& Qwen2.5 & 72B & 76.77 & 37.49 & 62.33 & 68.02 & 55.89 & 61.88 & 60.40 \\

& Qwen3 & 0.6B & 64.12 & 26.11 & 49.50 & 59.51 & 40.39 & 13.20 & 42.14 \\
& Qwen3 & 1.7B & 60.06 & 38.41 & 60.48 & 58.02 & 50.45 & 47.62 & 52.51 \\
& Qwen3 & 4B & 67.14 & 42.89 & 62.50 & 67.65 & 53.06 & 61.88 & 59.19 \\
& Qwen3 & 8B & 73.28 & 43.61 & 65.70 & 59.14 & 53.64 & 60.21 & 59.26 \\

& Llama-3.1 & 8B & 75.26 & 37.28 & 67.00 & 56.66 & 53.21 & 62.50 & 58.65 \\
& Llama-3.2 & 1B & 49.76 & 14.69 & 42.77 & 49.75 & 37.90 & 54.84 & 41.62 \\
& Llama-3.2 & 3B & 66.76 & 34.98 & 60.20 & 56.91 & 48.22 & 57.83 & 54.15 \\
& Llama-3.3 & 70B & \underline{81.87} & 38.16 & 63.62 & 66.42 & 55.50 & 65.58 & 61.86 \\

& Gemma-3 & 1B & 45.14 & 18.54 & 41.14 & 64.57 & 38.24 & 29.05 & 39.45 \\
& Gemma-3 & 4B & 61.95 & 28.16 & 57.34 & 62.47 & 40.17 & 48.42 & 49.75 \\
& Gemma-3 & 12B & 72.62 & 34.64 & 66.26 & 67.41 & 56.15 & 24.21 & 53.55 \\
& Gemma-3 & 27B & 80.36 & 35.23 & 67.10 & 65.56 & 58.93 & 45.95 & 58.85 \\

& Mistral-v0.1 & 7B & 69.69 & 25.56 & 54.32 & 57.16 & 47.07 & 58.98 & 52.13 \\
& Mistral-v0.2 & 7B & 69.41 & 32.47 & 57.51 & 64.07 & 50.06 & 59.77 & 55.55 \\
& Mistral-v0.3 & 7B & 72.33 & 29.12 & 56.67 & 63.33 & 51.61 & 60.04 & 55.52 \\

& DeepSeek-R1-Distill & 1.5B & 21.25 & 6.69 & 17.32 & 22.47 & 3.25 & 4.58 & 12.59 \\
& DeepSeek-R1-Distill & 7B & 27.38 & 7.15 & 22.25 & 13.21 & 5.05 & 55.37 & 21.74 \\
& DeepSeek-R1-Distill & 14B & 47.88 & 12.13 & 32.96 & 22.84 & 9.42 & 62.85 & 31.35 \\
& DeepSeek-R1-Distill & 32B & 35.03 & 10.00 & 26.46 & 21.73 & 7.07 & 63.12 & 27.23 \\
& DeepSeek-R1-Distill & 8B & 48.73 & 12.72 & 28.53 & 20.25 & 7.41 & 57.31 & 29.16 \\
& DeepSeek-R1-Distill & 70B & 43.63 & 11.38 & 27.13 & 14.32 & 4.71 & 65.05 & 27.70 \\
\hline

\multirow{4}{*}{\makecell[c]{Proprietary\\LLM}}
& DeepSeek-V4-Flash & 284B & 82.63 & 48.87 & 66.26 & \textbf{71.11} & \underline{58.42} & 67.69 & \underline{65.83} \\
& DeepSeek-V4-Pro & 1600B & 79.70 & 46.82 & 66.26 & 66.79 & 56.32 & 68.05 & 63.99 \\
& GPT-4.1 & 2200B$^{\dagger}$ & 81.30 & 46.82 & \underline{68.67} & 65.06 & 47.79 & \underline{69.19} & 63.14 \\
& GPT-5.1 & 1300B$^{\dagger}$ & 79.13 & \underline{49.92} & 67.66 & 62.59 & 45.52 & \textbf{72.89} & 62.95 \\
\hline

\multirow{3}{*}{\makecell[c]{Specialized\\LLM}}
& SusGen-GPT-Llama & 8B & 67.61 & 37.70 & 62.00 & 53.70 & 56.66 & 57.66 & 55.89 \\
& SusGen-GPT-Mistral & 7B & 71.20 & 34.10 & 58.52 & 62.84 & 52.98 & 61.44 & 56.85 \\
& \textsf{ESGPT}~(Ours) & 8B & \textbf{84.32~$\pm$~(0.09)} & \textbf{54.94~$\pm$~(0.16)} & \textbf{73.37~$\pm$~(0.31)} & 65.93~$\pm$~(0.17) & \textbf{59.40~$\pm$~(0.05)} & 63.82~$\pm$~(0.18) & \textbf{66.96~$\pm$~(0.10)}$^{*}$ \\
\hline

\end{tabular}%
}
\caption{Overall performance comparison on \textsf{ESGU-Bench}. The results are grouped by model type, with task-level scores reported across six ESG understanding categories. Best scores are shown in bold, second-best scores are underlined, and model sizes marked with $^{\dagger}$ indicate estimated parameter counts. To assess the performance stability of \textsf{ESGPT}, we report the mean $\pm$ standard deviation over five random seeds after excluding the highest and lowest seed results. The $^{*}$ marker in the Overall column denotes t-test significance against the second-best model, DeepSeek-V4-Flash, with $p < 0.001$.}
\label{tab:overall_performance}
\end{table*}
\begin{table*}[t]
\centering
\scriptsize
\setlength{\tabcolsep}{3.0pt}
\renewcommand{\arraystretch}{1.08}
\resizebox{\textwidth}{!}{%
\begin{tabular}{l|l|c|cccccc|c}
\hline
\textbf{Base Model} 
& \textbf{Model} 
& \textbf{Train Time}
& \makecell[c]{\textbf{ESG Relevance}\\\textbf{Detection}}
& \makecell[c]{\textbf{ESG Topic}\\\textbf{Identification}}
& \makecell[c]{\textbf{ESG Commitment}\\\textbf{Analysis}}
& \makecell[c]{\textbf{ESG Impact}\\\textbf{Analysis}}
& \makecell[c]{\textbf{ESG Claim}\\\textbf{Verification}}
& \makecell[c]{\textbf{ESG}\\\textbf{QA}}
& \textbf{Overall} \\
\hline

\multirow{5}{*}{\makecell[c]{Qwen3-8B}}
& ESGPT-10K & 03h 05m 49s & 81.68 & 53.72 & 69.28 & 63.58 & \underline{60.34} & 62.15 & 65.13 \\
& ESGPT-20K & 06h 41m 06s & 82.53 & 53.93 & 69.62 & 63.95 & \textbf{60.39} & 63.12 & 65.59 \\
& ESGPT-30K & 10h 16m 24s & \underline{84.32} & \underline{54.94} & \textbf{73.37} & \underline{65.93} & 59.40 & \underline{63.82} & \underline{66.96} \\
& ESGPT-50K & 17h 26m 58s & \textbf{84.80} & \textbf{55.69} & \underline{71.69} & \textbf{66.30} & 59.01 & \textbf{64.52} & \textbf{67.00} \\
& ESGPT-Full & 146h 29m 35s & 84.23 & 54.60 & 71.19 & 64.20 & 50.15 & 41.73 & 61.02 \\
\hline

\multirow{5}{*}{\makecell[c]{Llama-3.1-\\8B-Instruct}}
& ESGPT-10K & 03h 16m 39s & 81.78 & 49.21 & 69.06 & 61.36 & \underline{55.55} & 58.10 & 62.51 \\
& ESGPT-20K & 07h 20m 15s & 82.25 & 51.72 & 70.35 & 62.35 & \textbf{56.45} & 59.68 & 63.80 \\
& ESGPT-30K & 11h 23m 51s & 82.53 & \underline{52.72} & \textbf{71.97} & \underline{63.83} & 53.23 & \textbf{61.18} & \underline{64.24} \\
& ESGPT-50K & 19h 31m 03s & \textbf{84.14} & \textbf{52.80} & \underline{70.63} & \textbf{67.04} & 53.88 & \underline{59.86} & \textbf{64.72} \\
& ESGPT-Full & 164h 02m 13s & \underline{83.38} & 45.69 & 68.33 & 62.84 & 42.61 & 33.01 & 55.98 \\
\hline
\end{tabular}%
}
\caption{Scale ablation results of \textsf{ESGPT} on \textsf{ESGU-Bench}. The models are grouped by their base model, and each row reports the performance of a training data scale variant. Best scores within each base-model group are shown in bold.}
\label{tab:scale_ablation}
\end{table*}
\begin{table*}[t]
\centering
\scriptsize
\setlength{\tabcolsep}{3pt}
\renewcommand{\arraystretch}{1.12}
\begin{tabularx}{\textwidth}{@{}L{0.21\textwidth}L{0.12\textwidth}L{0.07\textwidth}Y L{0.05\textwidth}@{}}
\toprule
\textbf{Dataset} & \textbf{Source Type} & \textbf{Language} & \textbf{Description} & \textbf{Access} \\
\midrule

ESG-FTSE~\cite{pavlova2024esgftse}
& News / Media\newline Articles
& English
& English news dataset on top FTSE 100 firms for ESG relevance detection and primary E/S/G category classification.
& Public \\
\midrule

ESG-Kor~\cite{lee2024esgkor}
& Corporate Reports\newline / Disclosures
& Korean
& Korean ESG sentence dataset from sustainability reports, supporting ESG relevance detection and E/S/G category classification.
& Public \\
\midrule

ESGBERT~\cite{schimanski2024ESGBERT}
& Corporate Reports\newline / Disclosures
& English
& Sentence-level ESG corpus from corporate news, annual reports, and sustainability reports for E/S/G and action classification.
& Public \\
\midrule

ESG Activities~\cite{birti2025optimizing}
& Corporate Reports\newline / Disclosures
& English
& Transportation-sector non-financial disclosure dataset for matching text segments to EU ESG taxonomy activity descriptions.
& Public \\
\midrule

ClimateBERT~\cite{webersinke2021climatebert}
& Climate Reports\newline / Disclosures
& English
& Climate disclosure datasets for climate detection, specificity, sentiment, commitments/actions, TCFD, environmental claims, and target classification.
& Public \\
\midrule

DAX ESG Media~\cite{lipenkova2023detecting}
& News / Media\newline Articles
& English
& DAX company ESG media dataset using filtered external news and web documents for ESG topic extraction.
& Public \\
\midrule

DynamicESG~\cite{Tseng2023dynamicesg}
& News / Media\newline Articles
& Multilingual
& News-based ESG dataset for dynamic ESG category, impact type, and impact duration classification.
& Public \\
\midrule

A3CG~\cite{ong2025towards}
& Corporate Reports\newline / Disclosures
& English
& SGX sustainability report dataset for aspect extraction and action classification of corporate sustainability claims.
& Public \\
\midrule

PromiseEval~\cite{seki2025mlpromise}
& Corporate Reports\newline / Disclosures
& Multilingual
& Multilingual ESG report dataset for promise identification, supporting evidence, evidence clarity, and verification timing.
& Request \\
\midrule

SASB-Aligned ESG Sentences~\cite{junprung2023sasbesg}
& ESG Standards\newline / Frameworks
& English
& Sentence-level ESG dataset aligned with SASB standards for ESG detection and fine-grained SASB issue classification.
& Public \\
\midrule

ESG Sentiment~\cite{trajanovristo2023esgsentiment}
& Corporate Reports\newline / Disclosures
& English
& ESG aspect-level sentiment dataset with E/S/G-specific positive, neutral, and negative labels.
& Public \\
\midrule

ML-ESG-1~\cite{chen2023mlesg}
& News / Media\newline Articles
& Multilingual
& Multilingual ESG news dataset for identifying fine-grained ESG issues based on MSCI ESG key issue guidelines.
& Request \\
\midrule

ML-ESG-2~\cite{chen2023mlesg2}
& News / Media\newline Articles
& Multilingual
& Multilingual ESG dataset for classifying whether ESG-related events imply opportunity, risk, or cannot-distinguish impacts.
& Request \\
\midrule

ML-ESG-3~\cite{chen2024mlesg3}
& News / Media\newline Articles
& Multilingual
& Multilingual ESG news dataset for inferring ESG impact duration and related impact-level/type labels.
& Request \\
\midrule

SustainableQA~\cite{ali2025sustainableqa}
& Corporate Reports\newline / Disclosures
& English
& Large-scale sustainability QA dataset from corporate sustainability and annual reports, covering factoid, non-factoid, and table QA.
& Public \\
\midrule

ESGenius~\cite{he2025esgenius}
& ESG Standards\newline / Frameworks
& English
& Expert-validated ESG knowledge QA benchmark built from major sustainability standards and framework documents.
& Public \\
\midrule

Climate-FEVER~\cite{diggelmann2020climatefever}
& Wikipedia
& English
& Climate claim verification dataset linking real-world climate claims to Wikipedia evidence for support, refute, or NEI labels.
& Public \\

\bottomrule
\end{tabularx}
\caption{ESG resources used for constructing \textsf{ESG-Instruct} and \textsf{ESGU-Bench}.}
\label{tab:esg_resources}
\end{table*}
\newcolumntype{L}[1]{>{\RaggedRight\arraybackslash}p{#1}}
\newcolumntype{Y}{>{\RaggedRight\arraybackslash}X}
\newcolumntype{R}[1]{>{\RaggedLeft\arraybackslash}p{#1}}
\newcommand{\NA}{--}

\begin{table*}[t]
\centering
\scriptsize
\setlength{\tabcolsep}{2.5pt}
\renewcommand{\arraystretch}{1.08}
\begin{tabularx}{\textwidth}{
@{}
L{0.18\textwidth}
L{0.13\textwidth}
Y
L{0.065\textwidth}
R{0.065\textwidth}
R{0.065\textwidth}
R{0.065\textwidth}
R{0.065\textwidth}
@{}
}
\toprule
\textbf{Category} & \textbf{Dataset} & \textbf{Task} & \textbf{Language} & \textbf{Train} & \textbf{Valid} & \textbf{Test} & \textbf{Final} \\
\midrule

General
& Alpaca
& General Instruction Chat
& English
& 52000 & \NA & \NA & 3000 \\

\midrule
ESG Relevance Detection
& ESG-Kor
& Korean ESG Relevance Classification
& Korean
& 83262 & 16986 & 16987 & 5160 \\

\midrule
ESG Topic Identification
& A3CG
& Aspect Extraction
& English
& 1673 & 244 & 447 & 575 \\

& DAX ESG Media
& ESG Topic Classification
& English
& 11548 & \NA & \NA & 575 \\

& Dynamic-ESG
& ESG Category Classification
& Chinese
& 1058 & 118 & 131 & 575 \\

& ESG Activities
& ESG Activity--Concept Matching
& English
& 1272 & \NA & 53 & 575 \\

& ESG-BERT Dataset
& Environmental Classification
& English
& 2000 & \NA & \NA & 574 \\

& ESG-BERT Dataset
& Social Classification
& English
& 2000 & \NA & \NA & 574 \\

& ESG-BERT Dataset
& Governance Classification
& English
& 2000 & \NA & \NA & 574 \\

& ESG-Kor
& Korean ESG Category Classification
& Korean
& 79269 & 16986 & 16987 & 574 \\

& ML-ESG1
& ESG Issue Classification
& French
& 1200 & \NA & 300 & 574 \\

\midrule
ESG Commitment Analysis
& A3CG
& Action Classification
& English
& 2336 & 352 & 676 & 2688 \\

& ESG-BERT Dataset
& ESG Action Classification
& English
& 500 & \NA & \NA & 448 \\

& PromiseEval
& Promise Status Classification
& French
& 400 & \NA & 400 & 400 \\

& PromiseEval
& Promise Verification Timeline Classification
& French
& 400 & \NA & 400 & 400 \\

& PromiseEval
& Promise Status Classification
& Japanese
& 400 & \NA & 400 & 400 \\

& PromiseEval
& Promise Verification Timeline Classification
& Japanese
& 400 & \NA & 400 & 400 \\

\midrule
ESG Impact Analysis
& Dynamic-ESG
& Impact Duration Classification
& Chinese
& 995 & 140 & 156 & 470 \\

& Dynamic-ESG
& Impact Type Classification
& Chinese
& 1260 & 111 & 123 & 470 \\

& ESG-Sentiment
& ESG Sentiment Classification
& English
& 611 & \NA & 68 & 470 \\

& ML-ESG2
& Impact Type Classification
& French
& 818 & \NA & 200 & 470 \\

& ML-ESG2
& ESG Sentiment Classification
& Japanese
& 896 & \NA & 225 & 470 \\

& ML-ESG3
& Impact Duration Classification
& French
& 661 & \NA & 146 & 470 \\

& ML-ESG3
& Impact Level Classification
& French
& 661 & \NA & 146 & 470 \\

& ML-ESG3
& Impact Duration Classification
& Japanese
& 1500 & \NA & 53 & 470 \\

& ML-ESG3
& Impact Type Classification
& Japanese
& 1500 & \NA & 53 & 470 \\

& ML-ESG3
& Impact Duration Classification
& Korean
& 800 & \NA & 200 & 470 \\

& ML-ESG3
& Impact Type Classification
& Korean
& 800 & \NA & 200 & 470 \\

\midrule
ESG Claim Verification
& PromiseEval
& Evidence Quality Classification
& French
& 400 & \NA & 400 & 400 \\

& PromiseEval
& Evidence Status Classification
& French
& 400 & \NA & 400 & 400 \\

& PromiseEval
& Evidence Quality Classification
& Japanese
& 400 & \NA & 400 & 400 \\

& PromiseEval
& Evidence Status Classification
& Japanese
& 400 & \NA & 400 & 400 \\

\midrule
ESG Question Answering
& SustainableQA
& Factoid QA
& English
& 71418 & 8835 & 8609 & 1704 \\

& SustainableQA
& Non-Factoid QA
& English
& 82208 & 10140 & 10191 & 2066 \\

& SustainableQA
& Table QA
& English
& 3143 & 392 & 421 & 1394 \\

\bottomrule
\end{tabularx}
\caption{Composition of \textsf{ESG-Instruct}. Train/Valid/Test denote the number of instances in the original source dataset splits when available. Final denotes the number of selected instances used in \textsf{ESG-Instruct} after preprocessing and quality control.}
\label{tab:esg_instruct_composition}
\end{table*}

\begin{table*}[t]
\centering
\scriptsize
\setlength{\tabcolsep}{2.5pt}
\renewcommand{\arraystretch}{1.08}
\begin{tabularx}{\textwidth}{
@{}
L{0.18\textwidth}
L{0.13\textwidth}
Y
L{0.065\textwidth}
R{0.065\textwidth}
R{0.065\textwidth}
R{0.065\textwidth}
R{0.065\textwidth}
@{}
}
\toprule
\textbf{Category} & \textbf{Dataset} & \textbf{Task} & \textbf{Language} & \textbf{Train} & \textbf{Valid} & \textbf{Test} & \textbf{Final} \\
\midrule

ESG Relevance Detection
& ESG-FTSE
& ESG Relevance Classification
& English
& 3129 & 390 & 394 & 394 \\

& ClimateBERT
& Climate Detection Classification
& English
& 1300 & \NA & 400 & 400 \\

& ClimateBERT
& Environmental Claims Classification
& English
& 2117 & 265 & 265 & 265 \\

\midrule
ESG Topic Identification
& ESG-FTSE
& ESG Category Classification
& English
& 3127 & 389 & 396 & 396 \\

& ML-ESG1
& ESG Issue Classification
& English
& 1199 & \NA & 300 & 300 \\

& SASB-Aligned ESG Sentences
& SASB Parent Category Classification
& English
& 6460 & \NA & 647 & 647 \\

& SASB-Aligned ESG Sentences
& SASB Issue Topic Classification
& English
& 6460 & \NA & 647 & 647 \\

& ClimateBERT
& TCFD Recommendation Classification
& English
& 1300 & \NA & 400 & 400 \\

\midrule
ESG Commitment Analysis
& ClimateBERT
& Climate Commitments / Actions Classification
& English
& 1000 & \NA & 320 & 320 \\

& ClimateBERT
& Climate Specificity Classification
& English
& 1000 & \NA & 320 & 320 \\

& ClimateBERT
& Net-Zero Reduction Target Classification
& English
& 3441 & \NA & 344 & 344 \\

& PromiseEval
& Promise Status Classification
& English
& 400 & \NA & 400 & 400 \\

& PromiseEval
& Promise Verification Timeline Classification
& English
& 400 & \NA & 400 & 400 \\

\midrule
ESG Impact Analysis
& ClimateBERT
& Climate Sentiment Classification
& English
& 1000 & \NA & 320 & 320 \\

& ML-ESG2
& Impact Type Classification
& English
& 808 & \NA & 218 & 218 \\

& ML-ESG3
& Impact Duration Classification
& English
& 545 & \NA & 136 & 136 \\

& ML-ESG3
& Impact Level Classification
& English
& 545 & \NA & 136 & 136 \\

\midrule
ESG Claim Verification
& Climate Fever
& Claim Verification
& English
& \NA & \NA & 1535 & 1535 \\

& PromiseEval
& Evidence Quality Classification
& English
& \NA & \NA & 400 & 400 \\

& PromiseEval
& Evidence Status Classification
& English
& \NA & \NA & 400 & 400 \\

\midrule
ESG Question Answering
& ESGenius
& ESG Reasoning
& English
& \NA & \NA & 1136 & 1136 \\

\bottomrule
\end{tabularx}
\caption{Composition of \textsf{ESGU-Bench}. Train/Valid/Test denote the number of instances in the original source dataset splits when available. Final denotes the number of selected instances used in the \textsf{ESGU-Bench} evaluation set after preprocessing and quality control.}
\label{tab:esg_benchmark_composition}
\end{table*}

\end{document}